\documentclass{article}

\PassOptionsToPackage{numbers,sort&compress}{natbib}
\usepackage[final,main]{neurips_2026}

\usepackage{graphicx}
\usepackage[utf8]{inputenc}
\usepackage[T1]{fontenc}
\usepackage{hyperref}
\hypersetup{hidelinks}
\usepackage{url}
\usepackage{booktabs}
\usepackage{amsfonts}
\usepackage{amsmath}
\usepackage{nicefrac}
\usepackage{microtype}
\usepackage{xcolor}
\usepackage{multirow}
\usepackage{array}
\usepackage{fvextra}
\DefineVerbatimEnvironment{verbatim}{Verbatim}{breaklines=true,breakanywhere=true}

\usepackage[most]{tcolorbox}
\usepackage{needspace}
\definecolor{PromptTeal}{HTML}{508B77}
\definecolor{PromptBlue}{HTML}{668EBB}
\definecolor{PromptAmber}{HTML}{B98835}
\definecolor{PromptPurple}{HTML}{9582B9}
\newtcolorbox{promptcard}[3][]{
  enhanced,breakable,
  colback=#2!3!white,colframe=#2!80!black,
  colbacktitle=#2!14!white,coltitle=#2!35!black,
  title={#3},title after break={#3\hfill\normalfont\itshape continued},
  fonttitle=\normalfont\sffamily\bfseries\normalsize,
  boxrule=0.6pt,arc=1.5pt,
  left=9pt,right=9pt,top=7pt,bottom=7pt,
  toptitle=5pt,bottomtitle=5pt,
  before skip=9pt,after skip=12pt,
  pad at break*=5pt,#1
}

\newcommand{\pager}{\textsc{PAGER}}
\newcommand{\supp}{\textsc{Supported}}
\newcommand{\contra}{\textsc{Contradicted}}
\newcommand{\miss}{\textsc{Missing}}
\newcommand{\generic}{\textsc{Generic}}
\newcommand{\entity}{\textsc{Entity}}
\newcommand{\vdetail}{\textsc{Visual\_Detail}}
\newcommand{\interact}{\textsc{Interaction}}
\newcommand{\scene}{\textsc{Scene\_Context}}

\title{Where Do Long Captions Fail?\\
Position-Aware Diagnosis and Reinforcement Learning for Detailed Image Captioning}

\author{%
\parbox{\dimexpr\textwidth-2\tabcolsep\relax}{\centering
Yuanze Hu\textsuperscript{1,2,*}\quad
Zhichao Yang\textsuperscript{1}\quad
Junwei Jing\textsuperscript{2,3}\quad
Xin Yang\textsuperscript{2,4}\\[2pt]
Wenxuan Liu\textsuperscript{2}\quad
Tinghai Zhang\textsuperscript{2}\quad
Jia Fu\textsuperscript{2}\quad
Hongzhi Zhang\textsuperscript{2}\\[2pt]
Xiangyu Wu\textsuperscript{2,\dag}\quad
Ruiming Tang\textsuperscript{2}\quad
Tingting Gao\textsuperscript{2}\quad
Han Li\textsuperscript{2}\\[6pt]
\normalfont
\textsuperscript{1}Beihang University \quad
\textsuperscript{2}Kuaishou Technology\\[2pt]
\textsuperscript{3}Fudan University \quad
\textsuperscript{4}Zhejiang University
}}

\begin{document}

\maketitle
\begingroup
\renewcommand{\thefootnote}{\fnsymbol{footnote}}
\footnotetext[1]{Work done during an internship at Kuaishou Technology.}
\footnotetext[2]{Corresponding author.}
\endgroup

\begin{abstract}
Detailed image captioning is usually framed as an amount problem: a model should mention more visual facts while avoiding hallucination. However, autoregressive VLMs generate captions as ordered sequences, and later generations may be produced under weaker or noisier visual grounding. We therefore argue that long-form captioning should be studied not only by what facts are mentioned, but also by where they appear. We introduce a position-aware rubric representation that decomposes dense human references into atomic visual claims, verifies generated captions against those claims, and induces a quartile-wise composition of supported, hallucinated, and generic content. This lens reveals two qualitatively different failures in long captions: models tend to produce unsupported specificity in the middle, while the final portion often escapes into vague, non-evidential language rather than simply accumulating more hallucination. A same-reference human control does not exhibit these patterns; additional human audits examine the visual validity of the induced labels. Motivated by this diagnosis, we propose \pager{} (\textbf{P}osition-\textbf{A}ware \textbf{G}rounded \textbf{E}vidence \textbf{R}einforcement), a GRPO training objective with separate signals for supported fine-grained middle evidence, evidence-bearing tail continuation, claim coverage, and content-aware length control. The result is a closed loop: the same ordered claim representation identifies where long captions fail and supplies the position-specific reward needed to repair those failures. Experiments on detailed-caption benchmarks show that \pager{} improves position-wise grounding and pairwise caption preference while preserving competitive global caption quality.
\end{abstract}

\section{Introduction}

A detailed caption is not just a bag of visual facts. It is a sequence. The first sentences usually describe salient entities and global context; later sentences are expected to add attributes, relations, smaller objects, and scene-specific detail. Yet detailed-caption evaluation still largely reports global summaries: how many visual elements were covered, how many claims were factual, or which of two captions was preferred~\citep{dong2024benchmarking,ye2025painting,lu2025benchmarking,cheng2025caparena}. These scores are useful, but they collapse the temporal structure of generation. They can tell us that a long caption is imperfect, but not where it starts to fail or what kind of content replaces grounded description.

This paper starts from a simple question: \emph{when a model writes a long detailed caption, where do unsupported and uninformative sentences appear?} The question is motivated by recent evidence that multimodal generation can become less visually grounded as decoding proceeds~\citep{tang2025seeing,yang2025understanding,jung2025visual}. A mechanism-level decline in visual reliance, however, does not determine the semantic form of the failure. A model might hallucinate increasingly specific details; it might become vague; or it might alternate between the two. These cases require different training signals.

We answer the question with a position-aware rubric representation. For each image, a dense human reference is decomposed into atomic visual claims. A verifier judges whether each claim is supported, contradicted, or missing in a generated caption and returns the evidence span. By locating evidence spans inside the generated caption, we induce a quartile-wise composition $(S_q,H_q,G_q)$: supported, hallucinated, and generic content at each position. \generic{} is not produced by an auxiliary classifier; it is the residual category for sentences cited by no supported or contradicted claim. This distinguishes reference conflict from unmatched content, which can include filler and correct details absent from the reference.

Across six evaluated captioners from the Qwen, InternVL, and LLaVA families, this lens reveals two stable position-dependent failures. \textbf{Mid-caption unsupported specificity}: hallucination peaks in $Q_2/Q_3$, where models attempt to add fine-grained attributes and relations. \textbf{Tail generic escape}: the final quartile contains many more generic sentences than the first quartile. Both effects appear across the evaluated model families. Dense human references do not exhibit the patterns under a same-reference consistency check; independent image-based audits are reported in Appendix~\ref{app:image_audits}.

The central conceptual point is that these failures are not a single monotonic degradation. In three of six models, tail hallucination is lower than head hallucination even though tail genericity is much higher. The tail is not always more hallucinated; it is often less evidential. A vague sentence such as ``the overall atmosphere is calm and pleasant'' may be neither contradicted nor supported by any visual rubric. It avoids specific hallucination by escaping into generic language. A reward that simply encourages more late-caption content can therefore trade one failure for the other: it may reduce genericity by inducing fabricated detail. This motivates a split objective rather than a uniform late-position weight.

We propose \pager{} (\textbf{P}osition-\textbf{A}ware \textbf{G}rounded \textbf{E}vidence \textbf{R}einforcement), which turns the diagnostic representation into a GRPO reward. \pager{} combines claim coverage, a middle-position reward for supported fine-grained evidence, a tail-position penalty for generic escape, and a content-aware length budget. The reward components deliberately mirror the diagnostic findings: the middle and tail regions fail differently and are trained with different signals. The rubric data construction proceeds in two stages: we first build rubrics from curated dense-caption data, then train the final GRPO policy on a larger PixMo-derived rubric set filtered for image quality, caption length, and pre-inference judge quality.

Our contributions are:
\begin{enumerate}
    \item We introduce a position-aware rubric representation of detailed captions as ordered visual claims, enabling decomposition into supported, hallucinated, and generic content by caption position.
    \item We identify two reliable and qualitatively distinct position-dependent failures---mid-caption unsupported specificity and tail generic escape---and show that both are stable across six evaluated captioners but absent in dense human references.
    \item We distinguish tail substitution from monotonic error accumulation, challenging a simple ``later tokens are worse'' interpretation and directly constraining reward design.
    \item We propose \pager{}, a position-aware GRPO objective whose components map one-to-one onto the diagnostic findings, and evaluate whether it closes the loop by reducing the exact failures that motivated it.
\end{enumerate}

\section{A position-aware lens for long captions}
\label{sec:framework}

This section defines the object used throughout the paper. The goal is not to introduce another global caption metric. The goal is to build a common representation that supports both diagnostic measurement in Section~\ref{sec:diagnosis} and reward construction in Section~\ref{sec:pager}.

\paragraph{Reference rubrics.}
For an image $x$ paired with a dense human reference caption, an LLM extractor decomposes the reference into atomic visual claims
\begin{equation}
    \mathcal{R}(x)=\{r_i\}_{i=1}^{N_x}.
\end{equation}
Each rubric $r_i$ is visually checkable and has a dimension $d_i\in\{\entity,\vdetail,\interact,\scene\}$ and a weight $w_i$. We use default weights $1.0$ for \entity{}, $2.0$ for \vdetail{}, $1.5$ for \interact{}, and $1.0$ for \scene{}. Compound attributes referring to the same property of the same entity, such as ``black-and-white dog,'' are preserved as one claim to avoid artificial self-contradictions. Extraction prompts and normalization rules are in Appendix~\ref{app:pipeline}.

\paragraph{Position-aware verification.}
Given a generated caption $y$, a verifier judges each rubric against the full caption and returns
\begin{equation}
    z_i(y)\in\{\supp,\contra,\miss\}
\end{equation}
plus an evidence span quoted from $y$ whenever $z_i(y)\neq\miss$. We split $y$ into sentences $\{s_k\}_{k=1}^{n}$ and partition them into four ordered quartile bins by sentence index,
\begin{equation}
    Q_q=\bigl\{s_k\;:\;\bigl\lfloor 4(k-1)/n\bigr\rfloor+1=q\bigr\},\qquad q\in\{1,2,3,4\},
\end{equation}
so that $y=Q_1\cup Q_2\cup Q_3\cup Q_4$. A non-missing rubric inherits the quartile index $q_i(y)\in\{1,2,3,4\}$ of the sentence containing its evidence span, i.e.\ the unique $q$ with that sentence in $Q_q$. The resulting tuple $(r_i,z_i,q_i)$ is a position-aware claim judgment.

This verification is reference-supported rather than directly image-grounded. Mere absence from the reference does not imply contradiction: a sentence with no supporting or conflicting evidence remains \generic{}. An incorrect or incomplete reference can nevertheless create a false contradiction. We use this conservative setting because it gives a stable target across many images and captioners. Section~\ref{sec:discussion} and Appendices~\ref{app:limitations} and~\ref{app:image_audits} discuss and measure this limitation.

\paragraph{From claims to caption composition.}
The per-rubric judgments induce sentence labels without an auxiliary classifier. Let $\mathcal{R}_k^+$ be the supported rubrics whose evidence cites sentence $s_k$, and let $\mathcal{R}_k^-$ be the contradicted rubrics whose evidence cites $s_k$. We define
\begin{equation}
\ell(s_k)=
\begin{cases}
\supp & \text{if } \mathcal{R}_k^+\neq\emptyset,\\
\contra & \text{if } \mathcal{R}_k^+=\emptyset \text{ and } \mathcal{R}_k^-\neq\emptyset,\\
\generic & \text{if } \mathcal{R}_k^+=\emptyset \text{ and } \mathcal{R}_k^-=\emptyset.
\end{cases}
\end{equation}
A sentence is therefore \generic{} precisely when it contributes no evidence to any reference rubric, whether supported or contradicted. For each quartile bin $Q_q$, weighted sentence counts give the within-bin composition
\begin{equation}
    \bigl(S_{Q_q},H_{Q_q},G_{Q_q}\bigr),\qquad S_{Q_q}+H_{Q_q}+G_{Q_q}=1,
\end{equation}
where $S_{Q_q}$, $H_{Q_q}$, and $G_{Q_q}$ are the fractions of supported, contradicted, and generic sentence weight in $Q_q$, normalized by the total sentence weight $\sum_{s_k\in Q_q}\omega(s_k)$ within that bin (sentence weights $\omega(s_k)$ are defined in Appendix~\ref{app:pipeline}). Missing rubrics contribute to caption-level coverage but, having no evidence span, do not enter the within-bin denominator, which is why the three rates sum to one. The cross-model diagnosis uses rubric-dimension weights; Appendix~\ref{app:unweighted} reports its unweighted counterpart. Table~\ref{tab:main_position_results} reports the downstream policy comparison using unweighted sentence counts ($\omega(s_k)=1$).

\paragraph{Effect sizes.}
We summarize the two position effects, together with the matching shift in supported content, by
\begin{equation}
    D_{\mathrm{mid}}=\tfrac{1}{2}\bigl(H_{Q_2}+H_{Q_3}\bigr)-H_{Q_1},
    \qquad
    D_{\mathrm{tail}}=G_{Q_4}-G_{Q_1},
    \qquad
    \Delta S = S_{Q_4}-S_{Q_1}.
    \label{eq:effect_sizes}
\end{equation}
Positive $D_{\mathrm{mid}}$ means hallucination is elevated in the middle relative to the opening; positive $D_{\mathrm{tail}}$ means generic content is elevated in the final quartile; negative $\Delta S$ means supported content is depleted by the tail. We use these signed conventions consistently in Section~\ref{sec:diagnosis} and the main tables.

\paragraph{What distinguishes this from prior rubric work.}
Several recent methods already use rubrics, decomposed units, or checklist-style feedback for caption evaluation and training~\citep{ye2025painting,xing2025caprl,huang2026rubicap,tang2026cccaption}. We do not claim novelty for using rubrics per se. The distinction is threefold. First, our rubrics are \emph{position-aware}: each claim is tied to an evidence span and therefore to a region of the generated caption. Second, we induce \generic{} content as the residual of claim verification, which exposes a failure mode that hallucination-only metrics miss. Third, the reward is not a global rubric score; it targets the positions where the diagnostic shows distinct failures. This is why the same representation can support both a diagnostic claim and a training intervention.

\section{Diagnosing position-dependent failure}
\label{sec:diagnosis}

The diagnostic study asks whether long detailed captions exhibit systematic position-dependent failures, and whether those failures are artifacts of our pipeline or properties of model-generated text.

\begin{figure}[t]
\centering
\includegraphics[width=\linewidth]{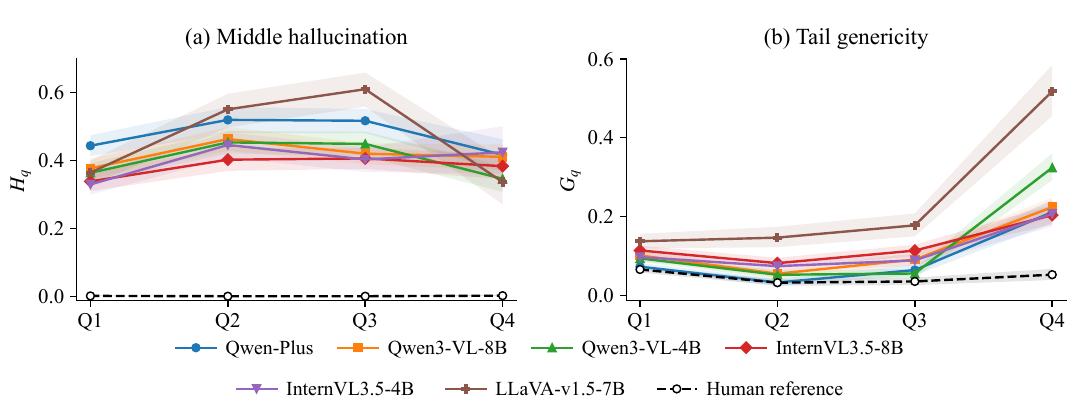}
\caption{\textbf{Position-dependent failures in detailed captioning.} (a) Hallucination rate $H_q$ peaks in the middle quartiles. (b) Generic rate $G_q$ rises in the final quartile. Colored solid lines denote six evaluated captioners; the dashed human line is a same-reference control. Statistics use rubric-dimension weights. Shaded bands show 95\% image-level paired bootstrap CIs.}
\label{fig:diagnostic}
\end{figure}

\paragraph{Setup.}
We evaluate six captioners on a mixed IIW-400 + DOCCI dense-caption evaluation set, using the subset of $n=198$ images for which predictions and reference rubrics are available. The evaluated models are Qwen-Plus, Qwen3-VL-8B/4B~\citep{bai2025qwen3}, InternVL3.5-8B/4B~\citep{wang2025internvl3}, and LLaVA-v1.5-7B~\citep{liu2023visual}. All verifier calls use deterministic decoding. For the cross-model diagnostic, uncertainty is estimated by image-level paired bootstrap with $B=2000$ replicates; Figure~\ref{fig:diagnostic} shows the 95\% confidence intervals. Verifier reliability is measured against human-majority labels in Appendix~\ref{app:verifier}. As a control, we apply the identical pipeline to dense human reference captions, using rubrics derived from each reference as that caption's verification target.

\paragraph{Mid-caption unsupported specificity.}
Figure~\ref{fig:diagnostic} shows that hallucination is not flat across the caption. For all six models, $H_q$ rises in the middle, and $D_{\mathrm{mid}}$ is positive with a 95\% CI excluding zero. The across-model mean is $+0.101$, with values ranging from $+0.065$ for Qwen3-VL-8B to $+0.216$ for LLaVA-v1.5-7B. We call this failure \emph{mid-caption unsupported specificity}. The head of a caption typically covers salient, easy-to-ground content; the middle is where models attempt to add finer attributes, smaller objects, and relations, and where many of these additions are unsupported by the reference. This reframes hallucination in long captions: the peak error region is where the caption is trying to become more informative.

\paragraph{Tail generic escape.}
The second pattern is different. Generic content rises sharply in the final quartile for every model. The weighted diagnostic gives mean $D_{\mathrm{tail}}=+0.178$, with model-level values from $+0.090$ for InternVL3.5-8B to $+0.380$ for LLaVA-v1.5-7B. Because generic sentences are defined as sentences cited by no rubric judgment, this effect means that the final quartile increasingly contains text unmatched to the reference-derived rubric. The corresponding supported content drops substantially: averaged across models, $\Delta S = S_{Q_4}-S_{Q_1}=-0.195$. Thus the end of the caption is not merely lower quality in a global sense; it is compositionally different.

\paragraph{Substitution rather than monotonic decay.}
\label{sec:substitution}
A tempting interpretation is that all errors worsen monotonically with position. The data reject that explanation. Comparing the tail to the head, three of the six models have lower hallucination in $Q_4$ than in $Q_1$: Qwen-Plus ($-0.027$), Qwen3-VL-4B ($-0.018$), and LLaVA-v1.5-7B ($-0.027$). The remaining three show modest increases from $+0.034$ to $+0.094$. Yet $D_{\mathrm{tail}}$ is large and positive for every model. The observed composition is consistent with substitution. Near the end of a long caption, models often avoid making concrete claims at all. A vague sentence may be neither contradicted nor supported; it simply does not serve as evidence for any visual claim. Hallucination-only metrics miss this behavior.

\begin{table}[t]
\centering
\small
\caption{Human-reference control. Applying the same pipeline to dense human references yields near-zero mid-caption hallucination and no tail generic escape. This is a same-reference consistency check rather than an independent completeness test.}
\begin{tabular}{lcc}
\toprule
Metric & Generated captions, mean of 6 models & Human references \\
\midrule
$S_{Q_1}$                  & 0.527       & 0.933 \\
$S_{Q_4}$                  & 0.333       & 0.945 \\
$G_{Q_1}$                  & 0.103       & 0.066 \\
$G_{Q_4}$                  & 0.281       & 0.053 \\
$D_{\mathrm{mid}}$ (CI95)  & $+0.101$    & $-0.001$ $[-0.003,+0.000]$ \\
$D_{\mathrm{tail}}$ (CI95) & $+0.178$    & $-0.013$ $[-0.028,+0.004]$ \\
\bottomrule
\end{tabular}
\label{tab:human_control}
\end{table}

\paragraph{Human-reference control.}
Table~\ref{tab:human_control} is a same-reference sanity check. Human references have weighted contradiction rates at or below $0.002$, $D_{\mathrm{mid}}\approx0$, and slightly negative $D_{\mathrm{tail}}$. Their tail support remains high ($S_{Q_4}=0.945$), whereas model captions average 0.333. Because the verification target is derived from the same text, this result does not test cross-writer agreement or reference under-coverage.

\begin{table}[t]
\centering
\small
\caption{The diagnostic findings translate into concrete constraints on reward design.}
\begin{tabular}{p{0.18\linewidth}p{0.36\linewidth}p{0.36\linewidth}}
\toprule
Requirement & Diagnostic reason & Reward implication \\
\midrule
Sub-caption granularity & Global scores hide where the composition changes. & Reward must operate on claims and sentence positions. \\
Mid-caption specificity & $H_{Q_2}/H_{Q_3}$ rises when models add fine details. & Reward supported fine-grained evidence in the middle, not arbitrary specificity. \\
Tail evidence density & $G_{Q_4}$ rises even when $H_{Q_4}$ does not. & Penalize non-evidential tail sentences without rewarding fabricated detail. \\
Length control & Coverage rewards can be hacked by verbosity. & Budget length according to image/reference complexity. \\
\bottomrule
\end{tabular}
\label{tab:requirements}
\end{table}

\paragraph{From diagnosis to reward design.}
For training, the implication is direct: reducing tail genericity by encouraging more specificity can create new tail hallucinations unless factuality and evidence density are controlled separately. Table~\ref{tab:requirements} summarizes the constraints that the diagnostic imposes on the reward. \pager{} implements one reward component per constraint.

\section{\pager{}: Position-Aware GRPO}
\label{sec:pager}

\begin{figure}[t]
\centering
\includegraphics[width=\linewidth]{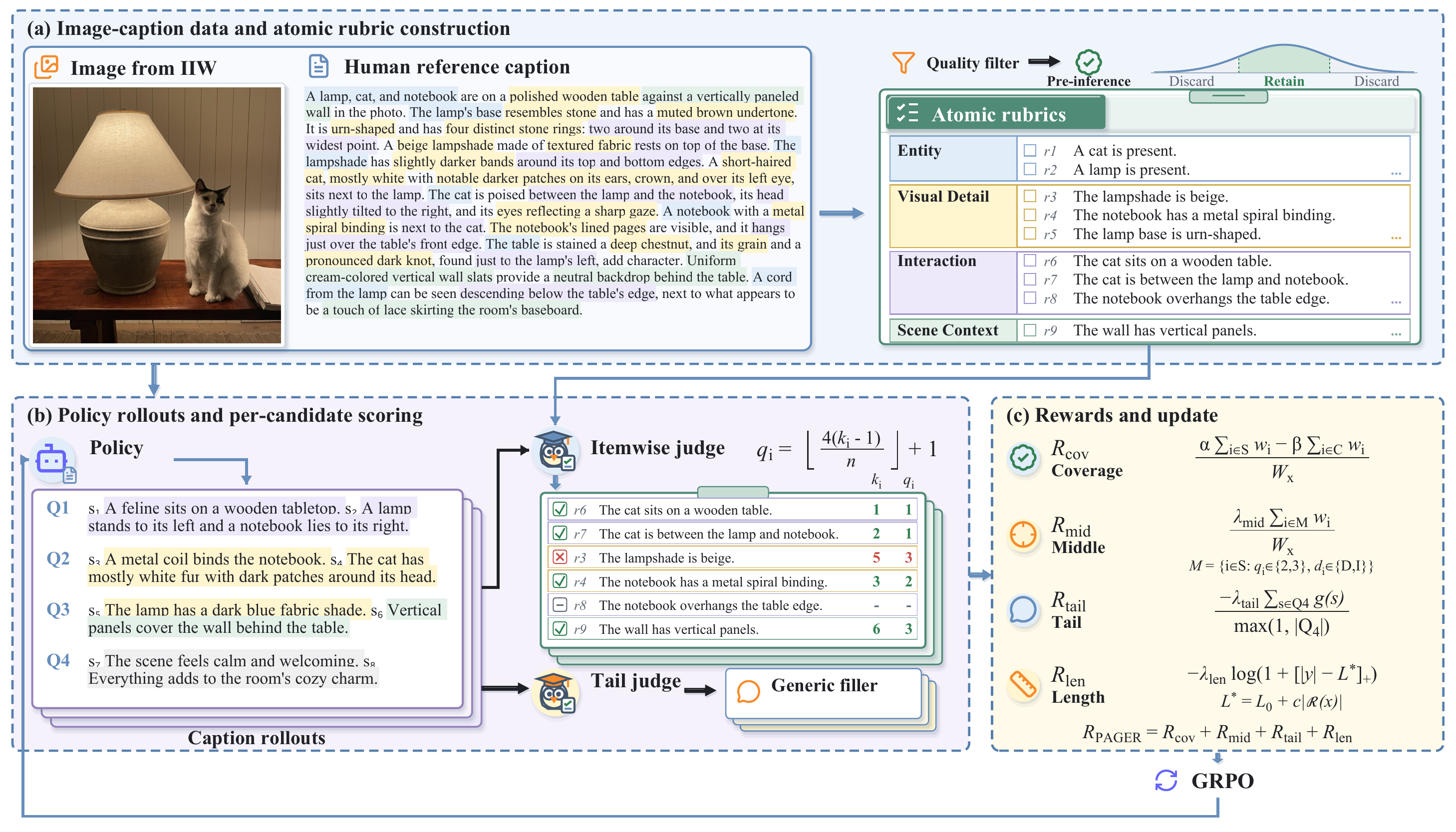}
\caption{\textbf{PAGER pipeline.} (a) Construct atomic rubrics from reference captions and filter training examples. (b) Sample policy captions and score each candidate with an itemwise rubric judge and a tail-genericity judge. (c) Combine four position-aware rewards to update the policy with GRPO.}
\label{fig:pipeline}
\end{figure}

\pager{} reuses the objects from Section~\ref{sec:framework}: rubrics $\mathcal{R}(x)$, itemwise verification labels $z_i(y)$, evidence quartiles $q_i(y)$, and induced sentence labels $\ell(s_k)$. Its reward is a sum of four terms, each matched to a diagnostic requirement (Figure~\ref{fig:pipeline}). The diagnostic, coverage reward, and mid-caption reward use itemwise checklist judgments with evidence spans; the tail reward uses a lightweight sentence-level genericity judge on final-quartile sentences. A holistic checklist judge, which returns only a single caption-level score, is used only as an ablation baseline and never for the position-wise analysis.

\paragraph{Claim coverage.}
The base factuality term rewards supported rubric items and penalizes contradicted items. Let
$W_x=\sum_{i=1}^{N_x}w_i$ be the total rubric weight for image $x$. We use the normalized reward
\begin{equation}
    R_{\mathrm{cov}}(y,x)=
    \frac{1}{W_x}
    \sum_{i=1}^{N_x} w_i\left[
    \alpha\,\mathbb{I}(z_i=\supp)-
    \beta\,\mathbb{I}(z_i=\contra)
    \right].
\end{equation}
Missing rubrics are neutral. This avoids forcing the model to guess details that it cannot ground and keeps reward magnitudes comparable across images with different numbers of rubrics.

\paragraph{Mid-caption specificity.}
The diagnostic shows that the middle failure is fabricated fine-grained detail. We therefore reward supported fine-grained rubrics whose evidence falls in $Q_2/Q_3$:
\begin{equation}
    R_{\mathrm{mid}}(y,x)=
    \lambda_{\mathrm{mid}}
    \frac{
    \sum_{i:\,z_i=\supp,\;q_i(y)\in\{2,3\},\;d_i\in\{\vdetail,\interact\}}
    w_i
    }{W_x}.
\end{equation}
Here $d_i$ is the dimension associated with rubric item $r_i$, and $q_i(y)$ is the caption quartile containing its evidence span. Rubric metadata supplies the claim type; evidence localization supplies its caption position. This term is restricted in both position and content type: it rewards reference-supported fine-grained middle evidence.

\paragraph{Tail evidence density.}
The tail failure is generic escape. During diagnosis, \generic{} is induced as the residual category of the itemwise verification pipeline. During training, to reduce reward latency and token cost, we use a sentence-level genericity judge only on final-quartile sentences. Let $g(s_k)=1$ if the sentence-level judge labels $s_k$ as generic and $0$ otherwise. A reconstructed reference prompt is provided in Appendix~\ref{app:tail_judge_prompt}. The tail reward is
\begin{equation}
    R_{\mathrm{tail}}(y,x)=
    -\lambda_{\mathrm{tail}}
    \frac{\sum_{s_k\in Q_4} g(s_k)}
    {\max\left(1,\left|\{s_k\in Q_4\}\right|\right)}.
\end{equation}
This term does not directly reward any non-generic tail sentence. If the model replaces vague filler with fabricated detail, the contradiction penalty in $R_{\mathrm{cov}}$ applies. The split between $R_{\mathrm{mid}}$ and $R_{\mathrm{tail}}$ is therefore essential: the two failure modes are structurally different.

\paragraph{Content-aware length budget.}
A coverage reward can be increased by writing longer captions. We set an image-specific budget
\begin{equation}
    L^\star(x)=L_0+c|\mathcal{R}(x)|
\end{equation}
for training examples with rubrics, and penalize only over-budget length:
\begin{equation}
    R_{\mathrm{len}}(y,x)=
    -\lambda_{\mathrm{len}}\log\left(1+\max(0,|y|-L^\star(x))\right).
\end{equation}
Here $|y|$ is computed as the character length of the generated caption in the implementation. At inference time, $|\mathcal{R}(x)|$ is unavailable; the reported evaluations therefore use the decoding budget described in Appendix~\ref{app:training}.

\paragraph{Rubric data curriculum.}
 We first used curated DCI and DOCCI training captions to build dense reference rubrics and validate the checklist pipeline. For the final GRPO training set, we moved to a larger PixMo-derived caption corpus. To avoid training on noisy rubrics, we filter examples using image-quality criteria, caption length constraints, and a pre-inference-and-judge pass: the current policy first generates candidate captions, the rubric judge scores their factuality and usefulness, and low-quality image--caption--rubric triples are removed. This curriculum preserves the clean structure of the original dense-caption rubrics while increasing diversity and scale for GRPO.

\paragraph{GRPO optimization.}
The full reward is
\begin{equation}
    R_{\pager}(y,x)=R_{\mathrm{cov}}+R_{\mathrm{mid}}+R_{\mathrm{tail}}+R_{\mathrm{len}}.
\end{equation}
We optimize with GRPO. For each image prompt, we sample a group of captions, compute $R_{\pager}$ for each caption, normalize rewards within the group to obtain relative advantages, and update the policy under a KL penalty to the reference policy. A schematic summary of the advantage and KL terms is
\begin{equation}
    \max_{\theta}\;\mathbb{E}_{y\sim\pi_\theta(\cdot|x)}
    \left[A(y,x)-\eta\,\mathrm{KL}(\pi_\theta\|\pi_{\mathrm{ref}})\right],
\end{equation}
where $A(y,x)$ is the group-relative advantage induced by $R_{\pager}$. Training hyperparameters are in Appendix~\ref{app:training}.

\section{Experiments}
\label{sec:experiments}

The experiments are organized around one central test: does \pager{} close the diagnostic loop? That is, after training, does the model reduce the mid-caption hallucination peak, sustain evidence-bearing content into the tail, and avoid replacing generic endings with new tail hallucinations?

\subsection{Setup}

\paragraph{Base model and training data.}
Base model: Qwen3-VL-4B-Instruct. \pager{} applies GRPO directly to this initialization; CapRL is treated as an external baseline when reported, not as the initialization for the main run. The reward rubrics were constructed through data filtering. Early rubric construction used curated DCI and DOCCI training captions; the final GRPO run uses a larger PixMo-derived caption corpus whose examples are filtered by image quality, caption length, pre-inference outputs, and judge scores to remove low-quality samples. Final main-run hyperparameters are reported in Appendix~\ref{app:training}; Appendix~\ref{app:construction_cost} gives the filtering funnel and measured token costs.

\paragraph{Baselines.}
The reported comparisons include: \textbf{Base}, the unmodified Qwen3-VL-4B-Instruct model; \textbf{SFT}, a supervised fine-tuning baseline on PixMo-Cap, reported as PixMol-4B; and \textbf{CapRL}~\citep{xing2025caprl}, treated as an external captioning/RL baseline rather than as our initialization. The ablations in Section~\ref{sec:experiments} isolate the value of \pager{}'s position-aware components by removing each reward term or replacing the split middle/tail design with simpler alternatives.

\paragraph{Benchmarks and metrics.}
We report global quality and density metrics on both IIW-400 and DOCCI, including CAPTURE, DCScore-F1, facts per 100 tokens, and average caption length. We also report four-model CapArena preference under the pairwise protocol~\citep{cheng2025caparena}. For each dataset, the CapArena value is computed only from the methods displayed in Table~\ref{tab:global_results}: each baseline is compared directly against \pager{}, ties are excluded from the win-rate denominator, and the \pager{} row pools the three direct comparisons. Table~\ref{tab:main_position_results} uses the same mixed IIW-400 + DOCCI diagnostic subset ($n=198$) as Section~\ref{sec:diagnosis}, measuring unweighted quartile-wise composition $(S_q,H_q,G_q)$. We report $\Delta S=S_{Q_4}-S_{Q_1}$ and $\Delta H=H_{Q_4}-H_{Q_1}$ separately to distinguish changes in tail support from changes in tail hallucination. Unless otherwise stated, all main results in Tables~\ref{tab:main_position_results} and~\ref{tab:global_results} are averaged over three repeated evaluation runs; the tables report the mean value. Repeated evaluations are distinct from independent training seeds; Appendix~\ref{app:scaling} separately reports the three-seed transfer update.

\subsection{Main results: global quality and positional repair}

\begin{table}[t]
\centering
\small
\setlength{\tabcolsep}{4pt}
\caption{Unweighted position metrics on the mixed IIW-400 + DOCCI diagnostic subset ($n=198$). $\Delta S=S_{Q_4}-S_{Q_1}$, $\Delta H=H_{Q_4}-H_{Q_1}$, and $D_{\mathrm{tail}}=G_{Q_4}-G_{Q_1}$. Best/second-best values are bold/underlined. Methods are defined in the setup.}
\begin{tabular}{lccccccc}
\toprule
Method & $D_{\mathrm{mid}}\downarrow$ & $S_{Q_4}\uparrow$ & $H_{Q_4}\downarrow$ & $G_{Q_4}\downarrow$ & $\Delta S\uparrow$ & $\Delta H\downarrow$ & $D_{\mathrm{tail}}\downarrow$ \\
\midrule
Base & +.096 & .290 & .298 & .412 & -.241 & -.028 & +.269 \\
SFT & \underline{+.091} & .296 & \underline{.267} & .437 & -.203 & \underline{-.053} & +.256 \\
CapRL & +.125 & \underline{.395} & .283 & \textbf{.322} & \underline{-.151} & +.028 & \textbf{+.123} \\
\textbf{\pager{}} & \textbf{+.036} & \textbf{.436} & \textbf{.235} & \underline{.329} & \textbf{-.094} & \textbf{-.083} & \underline{+.177} \\
\bottomrule
\end{tabular}
\label{tab:main_position_results}
\end{table}

\begin{table}[t]
\centering
\small
\setlength{\tabcolsep}{4pt}
\caption{Global benchmark summaries. CAPTURE and DCScore-F1 are quality metrics; facts per 100 tokens and mean token length describe density and verbosity. $\mathrm{CapArena}_4$ excludes ties: each baseline faces \pager{}, whose row pools those three comparisons. Scores are descriptive; no significance of small quality-score differences is claimed.}
\begin{tabular}{llccccc}
\toprule
Dataset & Method & CAPTURE$\uparrow$ & DCScore-F1$\uparrow$ & Facts/100tok$\uparrow$ & Length\,(tok) & $\mathrm{CapArena}_4\uparrow$ \\
\midrule
\multirow{4}{*}{IIW-400}
& Base & .638 & .507 & \underline{5.46}  & 257 & 4.32\%  \\
& SFT & .622 & .402 & \textbf{6.27} & 131 & 1.03\%  \\
& CapRL & \underline{.640} & \textbf{.552} & 3.20  & 475 & \underline{8.25\%}  \\
& \textbf{\pager{}} & \textbf{.642} & \underline{.526} & 3.43 & 412 & \textbf{96.28\%} \\
\midrule
\multirow{4}{*}{DOCCI}
& Base & .593 & .482 & \underline{5.32}  & 229 & 5.93\% \\
& SFT & .601 & .398 & \textbf{6.27}  & 128 & 2.06\% \\
& CapRL & \underline{.604} & \textbf{.524} & 3.28  & 464 & \underline{7.32\%} \\
& \textbf{\pager{}} & \textbf{.610} & \underline{.502} & 3.33  & 423 & \textbf{95.62\%} \\
\bottomrule
\end{tabular}
\label{tab:global_results}
\end{table}

Table~\ref{tab:main_position_results} is the central closed-loop result. On the same 198-image diagnostic subset, \pager{} reduces the mid-caption hallucination effect from $D_{\mathrm{mid}}=+.096$ for the base model to $+.036$, obtains the highest tail supported rate ($S_{Q_4}=.436$), and achieves the lowest tail hallucination rate ($H_{Q_4}=.235$). The tail repair is therefore not a trade-off where generic endings are replaced by fabricated details: relative to the base model, \pager{} both increases final-quartile supported content and lowers final-quartile hallucination. Table~\ref{tab:global_results} gives the complementary global picture. \pager{} has the best CAPTURE score on IIW-400 (.642) and DOCCI (.610), with small margins over CapRL. DCScore-F1 favors CapRL, so the evidence should be read as a trade-off: \pager{} improves CAPTURE and four-model pairwise preference while not maximizing every global metric. Facts density should also be interpreted with length in mind: the SFT and base models are denser partly because they produce much shorter captions, whereas \pager{} is denser than the longer CapRL baseline while producing shorter captions.

For CapArena, we compute a four-model tie-excluded pairwise win rate using the direct comparisons among the methods displayed in Table~\ref{tab:global_results}. For example, on IIW-400, \pager{} wins 133--6 against Qwen3-VL-4B-Instruct, 89--8 against CapRL-Qwen3VL-4B, and 192--2 against PixMol-4B, excluding ties. Pooling these comparisons gives $414/(414+16)=96.28\%$ for \pager{}.

\subsection{Ablations and robustness}

\begin{table}[t]
\centering
\small
\setlength{\tabcolsep}{3pt}
\caption{Component ablations under the position-aware diagnostic. Holistic checklist uses a caption-level score; monotone weight replaces separate middle/tail rewards with increasing late-position weight. Removing each reward term tests its contribution to the target region.}
\begin{tabular}{lcccccccc}
\toprule
Variant & $H_{Q_2}\downarrow$ & $H_{Q_3}\downarrow$ & $S_{Q_4}\uparrow$ & $H_{Q_4}\downarrow$ & $G_{Q_4}\downarrow$ & $D_{\mathrm{mid}}\downarrow$ & $\Delta S\uparrow$ & $\Delta H\downarrow$ \\
\midrule
\pager{} full
  & .364 & .344 & .436 & .235 & .329 & +.036 & -.094 & -.083 \\
w/o $R_{\mathrm{mid}}$
  & .421 & .408 & .418 & .241 & .341 & +.103 & -.112 & -.071 \\
w/o $R_{\mathrm{tail}}$
  & .371 & .350 & .374 & .248 & .378 & +.039 & -.156 & -.074 \\
w/o $R_{\mathrm{len}}$
  & .378 & .356 & .408 & .252 & .340 & +.047 & -.118 & -.068 \\
w/o filtering
  & .393 & .375 & .401 & .258 & .341 & +.064 & -.127 & -.062 \\
Holistic checklist
  & .388 & .367 & .421 & .244 & .335 & +.058 & -.103 & -.076 \\
Monotone weight
  & .370 & .347 & .402 & .291 & .307 & +.037 & -.138 & -.031 \\
\bottomrule
\end{tabular}
\label{tab:ablations}
\end{table}

\paragraph{Component ablations.}
Table~\ref{tab:ablations} supports the intended division of labor among the reward terms. Removing $R_{\mathrm{mid}}$ produces the clearest mid-caption degradation: $H_{Q_2}$ rises from .364 to .421, $H_{Q_3}$ from .344 to .408, and $D_{\mathrm{mid}}$ from +.036 to +.103. Removing $R_{\mathrm{tail}}$ mainly hurts the final quartile: $S_{Q_4}$ drops from .436 to .374 and $G_{Q_4}$ rises from .329 to .378, while $D_{\mathrm{mid}}$ remains close to the full model. Removing $R_{\mathrm{len}}$ and removing PixMo filtering both weaken the positional repair, increasing $D_{\mathrm{mid}}$ and lowering $S_{Q_4}$ relative to the full model. The holistic checklist reward is also weaker than the itemwise evidence-span reward, which suggests that caption-level supervision does not fully recover the targeted position-specific behavior. Finally, the monotone position-weight baseline has a similar mid-caption effect ($D_{\mathrm{mid}}=.037$ versus $.036$) and lowers $G_{Q_4}$ to .307 while raising $H_{Q_4}$ to .291, the highest tail hallucination rate among the ablations. This is the failure mode anticipated by the diagnostic: a single late-position incentive can reduce genericity by encouraging unsupported tail detail, whereas \pager{} separates middle specificity from tail evidence density.

\paragraph{Verifier reliability and human evaluation.}
Qwen3-8B supplies training judgments; DeepSeek-V3.2 remains outside the training loop. Both support the main positional comparison (Appendix~\ref{app:cross_judge}); their agreement with human labels on 347 rubric judgments is $\kappa=.89/.92$ (Appendix~\ref{app:verifier}). Supplementary image audits and human preferences provide descriptive evidence (Appendix~\ref{app:image_audits}--\ref{app:human_preference}). The post-RL residual-label audit describes the composition of unmatched tail sentences, with intervals spanning zero (Appendix~\ref{app:post_rl}).

\paragraph{Stronger models and transfer.}
Additional foundation-model diagnostics retain a positive tail effect (Appendix~\ref{app:foundation}). At a 100-step budget, three-seed 8B/32B training gives mean $D_{\mathrm{mid}}=.041/.043$ and $H_{Q_4}=.216/.214$. Appendix~\ref{app:scaling} presents single-run comparisons, CapRL continuation, and three-seed replication under a 100-step budget with per-scale learning rates.

\paragraph{Qualitative example.}
Appendix~\ref{app:qualitative} includes a concrete example for image \texttt{aar\_test\_04938}, comparing the base Qwen3-VL-4B-Instruct caption, the CapRL baseline, and \pager{}. The example illustrates how the positional patterns measured in Tables~\ref{tab:main_position_results}--\ref{tab:ablations} appear in an individual caption.

\section{Related work}
\label{sec:related}

\paragraph{Detailed captioning and evaluation.}
Image captioning has moved from short sentence generation toward dense, hyper-detailed description covering objects, attributes, relations, and scene context~\citep{onoe2024docci,urbanek2024picture,garg2024imageinwords,deitke2025molmo}. Classical metrics such as BLEU, METEOR, CIDEr, and SPICE measure overlap or scene-graph content~\citep{papineni-etal-2002-bleu,banerjee-lavie-2005-meteor,vedantam2015cider,anderson2016spice}. Recent detailed-caption metrics evaluate factuality, coverage, and preference: CAPTURE extracts visual elements~\citep{dong2024benchmarking}; DCScore decomposes captions into primitive units~\citep{ye2025painting}; CompreCap evaluates scene-graph completeness~\citep{lu2025benchmarking}; and CapArena and VELA evaluate long captions with preference- or long-caption-specific protocols~\citep{cheng2025caparena,matsuda2025vela}. These methods mostly produce global scores. Our work is complementary: we ask where supported, hallucinated, and generic content occurs inside the caption.

\paragraph{Rubrics, checklists, and feedback.}
Decomposed claims and checklists support both evaluation and optimization: FeedQuill uses DCScore-derived feedback~\citep{ye2025painting}; CapRL uses caption-only QA rewards~\citep{xing2025caprl}; RubiCap uses sample-specific rubrics~\citep{huang2026rubicap}; and CCCaption and SC-Captioner target correctness, completeness, and self-correction~\citep{tang2026cccaption,zhang2025sc}. \citet{huang2025rubricanchors} study rubric anchors for open-ended reinforcement learning. Beyond captioning, DeepSynth-Eval evaluates long-form survey synthesis through factual-coverage and structural-constraint checklists~\citep{zhang2026deepsynth}. PoSh uses scene-graph rubrics to localize detailed-caption errors, introduces DOCENT with expert references and human error annotations, and also serves as a reward~\citep{ananthram2026posh}. Our focus is position-dependent error composition and its implications for separate middle and tail rewards.

\paragraph{Hallucination and position effects.}
CHAIR, POPE, and FaithScore measure object- or claim-level visual hallucination~\citep{rohrbach2018object,li-etal-2023-evaluating,jing-etal-2024-faithscore}; FActScore verifies atomic claims in long-form text~\citep{min2023factscore}. At the representation level, TinyAlign improves cross-modal alignment through retrieved context~\citep{hu2026tinyalign}, while CoSE uses retrieved semantically connected latent samples to mitigate hallucinations~\citep{hu2026cose}. Recent studies suggest that multimodal models may rely less on visual information as generation proceeds~\citep{tang2025seeing,yang2025understanding,jung2025visual}, and long-context language models show position-dependent reliability~\citep{liu2023lostmiddle}. We provide a semantic, caption-level decomposition of such effects and show that late-caption failure is often generic escape rather than increased hallucination.

\section{Discussion and limitations}
\label{sec:discussion}

This paper argues that long detailed captions should be evaluated and trained as ordered sequences, rather than unordered collections of facts. Our diagnostic separates two failures that global scores conflate: unsupported specificity in the middle and generic escape at the tail. The latter is not merely ``more hallucination later,'' but a shift toward content unmatched to the reference rubric. \pager{} captures this structure by assigning separate reward terms to the corresponding failure regions.

\paragraph{Limitations.}
Reference verification can miss correct unreferenced details or create false contradictions. Additional human studies used unpaid volunteers from the authors' personal circle. The post-RL audit characterizes selected residual sentences descriptively. Diversity comparisons control sentence count while caption lengths differ. Transfer findings apply to the tested short-budget settings with per-scale learning rates; controlled scaling and reward-overfitting analyses require broader evaluation.

\paragraph{Broader implications.}
The same recipe---decomposing outputs into ordered claims, diagnosing position-wise composition, and aligning training signals to observed failure regions---may extend beyond captioning to long-form multimodal reasoning, text generation, and tool-use trajectories.

\begin{ack}
We thank the volunteers who helped with the human evaluations without monetary compensation. We thank Kuaishou Technology for providing computational resources and supporting the conference registration fee. Yuanze Hu carried out this work during an internship at Kuaishou Technology.
\end{ack}

\bibliographystyle{unsrtnat}
\bibliography{references}

\clearpage
\appendix
\raggedbottom

\section{Diagnostic pipeline details}
\label{app:pipeline}

\paragraph{Notation and visual conventions in Figure~\ref{fig:pipeline}.}
The example uses image \texttt{aar\_test\_04618} and its full IIW reference caption. Blue, yellow, purple, and green identify \entity{}, \vdetail{}, \interact{}, and \scene{}; gray marks generic filler. Rubric IDs and category colors are shared between construction and scoring. Check, cross, and dash denote \supp{}, \contra{}, and \miss{}. The schematic rollout has $n=8$ sentences $s_k$; $k_i$ is the one-based index of the evidence sentence for rubric $r_i$, and $q_i=\lfloor4(k_i-1)/n\rfloor+1$ is its quartile. Missing items have neither index. In the compact reward formulas, $S$ and $C$ are supported and contradicted rubric-index sets, while $M=\{i\in S:q_i\in\{2,3\},\ d_i\in\{D,I\}\}$, with $D/I$ denoting \vdetail{}/\interact{}. The notation $[t]_+=\max(0,t)$ abbreviates the length penalty. All other reward symbols are defined in Section~\ref{sec:pager}. The bell curve illustrates retaining intermediate pre-inference scores while excluding uninformative/unsolvable and already-saturated examples; it is not a fitted score distribution.

\subsection{Reference rubric construction}
For each image $x$, an LLM extractor decomposes the dense human reference caption into atomic visual rubrics. Each rubric item contains a claim, dimension, weight, and keyword. The extractor resolves pronouns to explicit nouns, preserves compound attributes referring to the same property of the same entity, deduplicates case- and whitespace-normalized claims, and discards empty or extremely short claims. The prompt is shown below.

\Needspace{8\baselineskip}
\begin{promptcard}{PromptTeal}{Prompt A.1 | Rubric extractor: system}
\begin{Verbatim}[fontsize=\normalsize,breaklines=true,breakanywhere=true,breakautoindent=true,breaksymbolleft={},baselinestretch=1]
You are an expert in Visual-Linguistic Decomposition.
Your task is to break down a ground-truth image caption into a structured list of atomic visual facts (a checklist).

### Objective
Create a high-precision verification set. We need to ensure the model captures both main subjects and fine-grained visual details without hallucination.

### Taxonomy & Weights
Classify each extracted fact into one of these dimensions:

1. Entity (Weight: 1.0): Core objects/nouns.
2. Visual_Detail (Weight: 2.0): Adjectives describing appearance (colors, materials, shapes, counts).
3. Interaction (Weight: 1.5): Relations between entities (spatial, actions).
4. Scene_Context (Weight: 1.0): Global context.

### Strict Constraints
1. Atomicity: One visual fact per item - but keep compound attributes together.
2. No Hallucination: Stick strictly to the input text. Do not infer or add information.
3. Coreference Resolution: Resolve pronouns like "it/they/them" to specific nouns.
4. Keyword: Pick the single most discriminative word as the anchor.
5. Format: Output valid JSON only.
6. Compound Attributes Rule (CRITICAL):
   - When multiple values describe the SAME property of an entity, keep them in ONE claim.
   - Properties include: color, material, size, pattern, texture, etc.

   CORRECT Examples:
   - "white and black dog" -> "The dog has white and black coloring"
   - "wooden and metal table" -> "The table is made of wood and metal"
   - "large and round eyes" -> "The eyes are large and round"
   - "red, blue and yellow stripes" -> "The stripes are red, blue and yellow"

   WRONG Examples (DO NOT do this):
   - "white and black dog" -> Split into "The dog is white" + "The dog is black"
   - "wooden and metal table" -> Split into "The table is wooden" + "The table is metal"

7. Semantic Completeness: Each claim must be semantically complete and non-contradictory when evaluated independently.

### Decision Tree for Compound Attributes
Ask yourself: "If I split this into separate claims, would they contradict each other or lose meaning?"
- If YES -> Keep together in one claim
- If NO -> Can split into separate atomic facts

Examples:
- "A red car and a blue truck" -> Split OK (different entities)
- "A red and blue car" -> Keep together (same entity, same property)
- "The dog is sitting and wagging its tail" -> Split OK (different actions)
- "The ball is large and round" -> Keep together (related size/shape of same object)
\end{Verbatim}
\end{promptcard}

\Needspace{8\baselineskip}
\begin{promptcard}[breakable=false]{PromptTeal}{Prompt A.2 | Rubric extractor: user template}
\begin{Verbatim}[fontsize=\normalsize,breaklines=true,breakanywhere=true,breakautoindent=true,breaksymbolleft={},baselinestretch=1]
# Task
Decompose the following text into atomic visual facts following the rules in the system prompt.

# Input Text
"""{gt_text}"""

# Critical Reminder
- Compound Attributes: Keep multiple values for the SAME property together
  - "white and black dog" -> ONE claim: "The dog has white and black coloring"
  - NOT two claims: "The dog is white" + "The dog is black"

# Output Schema (JSON)
{
  "rubric": [
    {
      "claim": "Atomic sentence with resolved nouns and compound attributes preserved.",
      "dimension": "Entity|Visual_Detail|Interaction|Scene_Context",
      "weight": <float>,
      "keyword": "most_discriminative_anchor_word"
    }
  ]
}

# Output Requirements
- Valid JSON format
- Each claim must be verifiable from the input text
- Preserve compound attributes (colors, materials, sizes) in single claims
- Resolve all pronouns to specific nouns
\end{Verbatim}
\end{promptcard}

\subsection{Position-aware checklist verification}
Given generated caption $y$ and rubric set $\mathcal{R}(x)$, the main judge is itemwise: it returns a status $z_i\in\{\supp,\contra,\miss\}$ and evidence span for every rubric. The judge sees the full caption, not presegmented sentences. We independently segment $y$ into sentences using punctuation-based splitting, discard very short fragments, and assign quartiles by sentence index. For a caption with $n$ valid sentences indexed $k=1,\dots,n$, sentence $s_k$ is assigned to the quartile bin $Q_q$ with index
\begin{equation}
    q_k=\left\lfloor\frac{4(k-1)}{\max(n,1)}\right\rfloor+1\in\{1,2,3,4\},
\end{equation}
matching the partition $Q_q=\{s_k:q_k=q\}$ used in Section~\ref{sec:framework}.

Evidence spans are matched back to sentences by exact string matching after normalization; unmatched spans are handled by fuzzy matching to the nearest sentence, and examples with unresolved evidence are excluded from the position-wise aggregation.

\Needspace{24\baselineskip}
The following itemwise prompt is used for the main diagnostic and \pager{} reward.
\Needspace{8\baselineskip}
\begin{promptcard}[breakable=false]{PromptBlue}{Prompt A.3 | Itemwise rubric judge}
\begin{Verbatim}[fontsize=\normalsize,breaklines=true,breakanywhere=true,breakautoindent=true,breaksymbolleft={},baselinestretch=1]
You are a strict fact-checking assistant.
Given a CAPTION and a list of CHECKLIST ITEMS, evaluate the relationship.
For each item, classify the status as one of the following:
- 'supported': The caption explicitly supports the claim.
- 'contradicted': The caption explicitly contradicts the claim or states the opposite.
- 'missing': The caption does not mention the claim at all.
Return ONLY a JSON object where keys are the item indices (0, 1, 2...) and values are objects with `status` (string) and `evidence` (string, leave empty if missing).
Caption: {caption}

Checklist Items:
{checklist_text}

JSON Output:
\end{Verbatim}
\end{promptcard}

\Needspace{24\baselineskip}
\paragraph{Sentence-level tail-genericity judge: reference prompt.}
\label{app:tail_judge_prompt}
The following reference template documents the sentence-level decision rule used in training; its wording is reconstructed for documentation. It classifies a sentence's specificity rather than verifying its truth against an image or reference.
\Needspace{8\baselineskip}
\begin{promptcard}[breakable=false]{PromptAmber}{Prompt A.4 | Tail-genericity judge}
\begin{Verbatim}[fontsize=\normalsize,breaklines=true,breakanywhere=true,breakautoindent=true,breaksymbolleft={},baselinestretch=1]
Decide whether this sentence is generic filler in an image caption.
Set generic=1 if it contains only vague summaries, subjective impressions, or commentary, with no concrete visually checkable detail.
Set generic=0 if it contains a concrete object, attribute, action, or spatial relation. Specificity alone does not establish factual correctness.
Return only JSON: {"generic": 0} or {"generic": 1}.
Sentence: {sentence}
\end{Verbatim}
\end{promptcard}
The returned binary value is $g(s_k)$ in the tail-reward definition. Apply it only to $s_k\in Q_4$ and average the flags with the negative coefficient $-\lambda_{\mathrm{tail}}$. Diagnostic $G_q$ remains the residual category from itemwise verification.

\Needspace{23\baselineskip}
\paragraph{Holistic judge prompt used only for ablation.}
The holistic judge removes itemwise evidence spans and therefore cannot produce the position-aware quantities $(S_q,H_q,G_q)$. We include it only as an ablation to test whether a single caption-level checklist score can substitute for itemwise position-aware supervision.
\Needspace{8\baselineskip}
\begin{promptcard}[breakable=false]{PromptPurple}{Prompt A.5 | Holistic judge: ablation only}
\begin{Verbatim}[fontsize=\normalsize,breaklines=true,breakanywhere=true,breakautoindent=true,breaksymbolleft={},baselinestretch=1]
You are a strict evaluation assistant.
Given a CAPTION and a list of CHECKLIST ITEMS, evaluate how well the caption satisfies the checklist as a whole.
Consider completeness, factual consistency, and overall coverage.
Return ONLY a JSON object with:
- `score`: a float between 0.0 and 1.0
- `evidence`: a brief explanation string
Caption: {caption}

Checklist Items:
{checklist_text}

JSON Output:
\end{Verbatim}
\end{promptcard}

\subsection{Sentence-level induced labels and aggregation}
For each sentence $s_k$, let $\mathcal{R}_k^+$ and $\mathcal{R}_k^-$ denote the sets of supported and contradicted rubrics whose evidence span cites $s_k$. The induced sentence label $\ell(s_k)\in\{\supp,\contra,\generic\}$ follows the rule in Section~\ref{sec:framework}. Each sentence receives a weight
\begin{equation}
\omega(s_k)=
\begin{cases}
\sum_{r_i\in\mathcal{R}_k^+}w_i & \ell(s_k)=\supp,\\
\sum_{r_i\in\mathcal{R}_k^-}w_i & \ell(s_k)=\contra,\\
1 & \ell(s_k)=\generic.
\end{cases}
\end{equation}
Let $W_{Q_q}=\sum_{s_k\in Q_q}\omega(s_k)$ denote the total sentence weight in quartile bin $Q_q$ (defined in Section~\ref{sec:framework}). Under the correspondence $c(S)=\supp$, $c(H)=\contra$, $c(G)=\generic$, the within-bin rates are
\begin{equation}
X_{Q_q}=\frac{1}{W_{Q_q}}\sum_{s_k\in Q_q}\omega(s_k)\,\mathbb{I}\bigl[\ell(s_k)=c(X)\bigr],
\qquad X\in\{S,H,G\}.
\end{equation}
Missing rubrics contribute to caption-level coverage but, having no evidence span, do not enter any bin's denominator $W_{Q_q}$; consequently $S_{Q_q}+H_{Q_q}+G_{Q_q}=1$ within every bin.

\subsection{Bootstrap confidence intervals}
Confidence intervals for the cross-model diagnostic rates and effects use image-level paired bootstrap with $B=2000$ replicates. Each replicate samples images with replacement, pools all sentences from sampled images, and re-aggregates quartile counts. This accounts for correlation among sentences from the same image.

\section{Additional diagnostic statistics}
\label{app:diagnostic_stats}

\begin{table}[htbp]
\centering\small
\setlength{\tabcolsep}{4pt}
\caption{Weighted per-model rubric-based position-wise composition on the mixed IIW-400 + DOCCI set ($n=198$). S/H/G are weighted sentence-level rates. Effect sizes use image-level paired bootstrap ($B=2000$).}
\begin{tabular}{lcccc}
\toprule
Model & $Q_1$ S/H/G & $Q_2$ S/H/G & $Q_3$ S/H/G & $Q_4$ S/H/G \\
\midrule
Qwen-Plus & .483/.444/.073 & .447/.520/.033 & .418/.517/.065 & .372/.417/.211 \\
Qwen3-VL-8B & .521/.377/.102 & .481/.463/.055 & .488/.420/.091 & .365/.411/.224 \\
Qwen3-VL-4B & .542/.364/.094 & .494/.453/.053 & .495/.449/.056 & .330/.346/.324 \\
InternVL3.5-8B & .547/.339/.114 & .515/.403/.082 & .480/.406/.114 & .413/.383/.204 \\
InternVL3.5-4B & .573/.330/.097 & .480/.446/.075 & .508/.403/.089 & .369/.424/.207 \\
LLaVA-v1.5-7B & .498/.364/.138 & .302/.551/.147 & .212/.610/.178 & .146/.337/.517 \\
Human ref. & .933/.001/.066 & .967/.000/.033 & .964/.000/.036 & .945/.002/.053 \\
\bottomrule
\end{tabular}
\medskip

\begin{tabular}{lcc}
\toprule
Model & $D_{\mathrm{mid}}$ [95\% CI] & $D_{\mathrm{tail}}$ [95\% CI] \\
\midrule
Qwen-Plus & +.075 [+.032,+.116] & +.138 [+.110,+.169] \\
Qwen3-VL-8B & +.065 [+.036,+.093] & +.122 [+.098,+.149] \\
Qwen3-VL-4B & +.087 [+.055,+.120] & +.230 [+.199,+.264] \\
InternVL3.5-8B & +.066 [+.032,+.100] & +.090 [+.063,+.118] \\
InternVL3.5-4B & +.095 [+.060,+.127] & +.110 [+.077,+.143] \\
LLaVA-v1.5-7B & +.216 [+.162,+.263] & +.380 [+.315,+.450] \\
Human ref. & $-$.001 [$-$.003,+.000] & $-$.013 [$-$.028,+.004] \\
\bottomrule
\end{tabular}
\label{tab:appendix_position_stats_weighted}
\end{table}

\subsection{Unweighted diagnostic statistics}
\label{app:unweighted}

\begin{table}[htbp]
\centering\small
\setlength{\tabcolsep}{4pt}
\caption{Unweighted position-wise diagnostic statistics on the mixed IIW-400 + DOCCI subset ($n=198$). This table reports the same diagnostic without rubric-dimension weighting, supporting that the two position effects are not artifacts of the weighting scheme. $\Delta_{\mathrm{halluc}}=H_{Q_4}-H_{Q_1}$ and $\Delta_{\mathrm{supported}}=S_{Q_1}-S_{Q_4}$.}
\begin{tabular}{lcccc}
\toprule
Model & $Q_1$ S/H/G & $Q_2$ S/H/G & $Q_3$ S/H/G & $Q_4$ S/H/G \\
\midrule
InternVL3.5-4B & .562/.289/.149 & .477/.407/.116 & .504/.366/.131 & .351/.362/.287 \\
InternVL3.5-8B & .529/.297/.174 & .498/.374/.128 & .459/.370/.171 & .367/.343/.291 \\
Qwen3-VL-4B & .531/.326/.142 & .496/.422/.082 & .497/.421/.082 & .290/.298/.412 \\
Qwen3-VL-8B & .515/.329/.156 & .479/.432/.089 & .474/.389/.137 & .335/.347/.318 \\
LLaVA-v1.5-7B & .486/.336/.178 & .285/.511/.204 & .200/.573/.227 & .120/.293/.586 \\
Qwen-Plus & .487/.398/.115 & .462/.485/.053 & .426/.476/.098 & .359/.349/.293 \\
Human ref. & .899/.001/.100 & .943/.000/.057 & .943/.000/.057 & .921/.001/.078 \\
\bottomrule
\end{tabular}
\medskip

\begin{tabular}{lccccc}
\toprule
Model & Units & $D_{\mathrm{mid}}$ & $D_{\mathrm{tail}}$ & $\Delta H$ & $S_{Q_1}-S_{Q_4}$ \\
\midrule
InternVL3.5-4B & 4744 & +.098 & +.138 & +.073 & +.211 \\
InternVL3.5-8B & 5353 & +.075 & +.117 & +.046 & +.162 \\
Qwen3-VL-4B & 5623 & +.096 & +.270 & -.028 & +.241 \\
Qwen3-VL-8B & 6339 & +.082 & +.162 & +.018 & +.180 \\
LLaVA-v1.5-7B & 2130 & +.206 & +.408 & -.043 & +.365 \\
Qwen-Plus & 4894 & +.083 & +.178 & -.049 & +.129 \\
Human ref. & 4582 & -.001 & -.022 & +.000 & -.022 \\
\bottomrule
\end{tabular}
\label{tab:unweighted_position_stats}
\end{table}

The main diagnostic aggregates the mixed IIW-400 + DOCCI evaluation subset of 198 images to characterize position effects across both dense-caption sources.

\section{Verifier reliability}
\label{app:verifier}

We assess the verifier on 347 held-out rubric--caption judgments stratified by model and quartile. Each rubric-level judgment is labeled by human annotators as \supp{}, \contra{}, or \miss{}, and we compare three LLM judges to the human-majority label. Inter-annotator agreement is Cohen's $\kappa=0.94$. We use human-majority labels as the reliability target.

\begin{table}[h]
\centering
\small
\caption{Verifier reliability against human annotations. All LLM judges are within $0.05$ of the human inter-annotator ceiling in Cohen's $\kappa$; the contradicted class remains the hardest.}
\begin{tabular}{lcccc}
\toprule
Source & $\kappa$ vs. human & F1-Supp. & F1-Contr. & F1-Miss. \\
\midrule
Primary judge (Qwen3-8B)          & 0.89 & 0.92 & 0.87 & 0.91 \\
Secondary judge (Qwen3-14B)       & 0.90 & 0.92 & 0.88 & 0.91 \\
Independent judge (DeepSeek-V3.2) & 0.92 & 0.94 & 0.90 & 0.93 \\
Inter-annotator ceiling           & 0.94 & 0.96 & 0.92 & 0.95 \\
\bottomrule
\end{tabular}
\label{tab:verifier_reliability}
\end{table}

The human-reference control in Table~\ref{tab:human_control} provides an additional sanity check: when the verifier evaluates dense human captions against rubrics derived from those captions, reference contradiction remains at or below $0.002$ in every quartile.

\section{Training details}
\label{app:training}

\paragraph{Model and data.}
The main GRPO policy is initialized directly from Qwen3-VL-4B-Instruct. The final GRPO prompts come from a PixMo-derived long-caption corpus stored as a parquet dataset. During training, validation uses the DCI densely-captioned-images test parquet. The prompt key is \texttt{problem}, the answer key is \texttt{answer}, and images are read from \texttt{images}. Earlier rubric-construction experiments used DCI and DOCCI training captions, but the reported main GRPO run uses the filtered PixMo-derived rubric data described in Section~\ref{sec:pager}.

\paragraph{Run configuration.}
Table~\ref{tab:training_hparams} reports the main-run configuration. The main GRPO run used 4 nodes with 8 A100 GPUs per node and took approximately two days of wall-clock time.

\begin{table}[h]
\centering
\small
\caption{Main GRPO training hyperparameters. }
\begin{tabular}{p{0.38\linewidth}p{0.54\linewidth}}
\toprule
Setting & Value \\
\midrule
Base model & Qwen3-VL-4B-Instruct \\
Training data & filtered PixMo-derived long-caption parquet \\
Training validation file & DCI densely-captioned-images test parquet \\
Prompt / answer / image keys & \texttt{problem} / \texttt{answer} / \texttt{images} \\
Prompt template & caption chat template \\
Shuffle / seed & true / 1 \\
Filter overlong prompts & disabled \\
Maximum prompt length & 6144 tokens \\
Maximum response length & 8192 tokens \\
Minimum / maximum image pixels & 262{,}144 / 4{,}194{,}304 \\
Rollout group size & 8 captions per prompt \\
Rollout temperature / top-$p$ & 1.0 / 1.0 \\
Validation decoding & temperature 0.6, top-$p$ 0.95, $n=1$ \\
Rollout batch size & 1024 \\
Validation batch size & 1024 \\
Actor global batch size & 256 \\
Actor update micro-batch per device & 2 \\
Experience/log-prob micro-batch per device & 4 \\
Maximum batched rollout tokens & 14{,}336 \\
Epochs & 3 \\
Nodes / GPUs per node & 4 / 8 \\
\bottomrule
\end{tabular}
\label{tab:training_hparams}
\end{table}

\paragraph{Optimization and KL control.}
We optimize the Qwen3-VL-4B-Instruct policy with GRPO and KL regularization to the reference policy. Table~\ref{tab:optimization_hparams} reports the optimizer and systems settings used for the main run. The reward implementation is the remote itemwise checklist reward described in Section~\ref{sec:pager}. Table~\ref{tab:reward_hparams} reports the reward hyperparameters used by the main run.

\begin{table}[h]
\centering
\small
\caption{Optimization and systems configuration for the main GRPO run.}
\begin{tabular}{p{0.38\linewidth}p{0.54\linewidth}}
\toprule
Setting & Value \\
\midrule
Advantage estimator & GRPO \\
KL loss / penalty & enabled; low-variance KL \\
KL coefficient & $1.0\times10^{-2}$ \\
Optimizer & AdamW \\
Learning rate & $1.0\times10^{-6}$ \\
Weight decay & $1.0\times10^{-2}$ \\
Warmup ratio & 0.0 \\
Maximum gradient norm & 1.0 \\
Online filtering & disabled \\
Padding-free / dynamic batching & enabled / enabled \\
Ulysses size & 1 \\
LoRA & disabled (rank 0); visual-module exclusion pattern specified but inactive \\
Gradient checkpointing & enabled \\
Vision tower & not frozen \\
Actor FSDP & full shard enabled, CPU offload disabled, rank-0 init enabled \\
Actor offload & parameter offload disabled; optimizer offload enabled \\
Reference FSDP & full shard enabled, CPU offload disabled, rank-0 init enabled \\
Reference offload & parameter offload disabled \\
Rollout backend settings & vLLM V1 enabled, tensor parallel size 1, GPU memory utilization 0.9, eager mode disabled, chunked prefill disabled \\
Data/system settings & DataLoader workers 2; Ray object store 10GB; plasma directory on local disk; Arrow IPC zero-copy disabled \\
\bottomrule
\end{tabular}
\label{tab:optimization_hparams}
\end{table}

\begin{table}[h]
\centering
\small
\caption{Reward hyperparameters for the main \pager{} run. The reward is $R_{\pager}=R_{\mathrm{cov}}+R_{\mathrm{mid}}+R_{\mathrm{tail}}+R_{\mathrm{len}}$.}
\begin{tabular}{p{0.30\linewidth}p{0.23\linewidth}p{0.36\linewidth}}
\toprule
Parameter & Value & Role \\
\midrule
$\alpha$ & 1.0 & Reward for a supported checklist item in $R_{\mathrm{cov}}$ \\
$\beta$ & 0.5 & Penalty weight for a contradicted checklist item in $R_{\mathrm{cov}}$ \\
$\lambda_{\mathrm{mid}}$ & 0.3 & Weight for supported fine-grained evidence in $Q_2/Q_3$ \\
$\lambda_{\mathrm{tail}}$ & 0.2 & Penalty weight for generic final-quartile sentences \\
$L_0$ & 512 & Base caption length budget \\
$c$ & 100 & Additional length budget per rubric item \\
$\lambda_{\mathrm{len}}$ & 0.05 & Weight for the logarithmic over-budget length penalty \\
Judge model & \texttt{models/Qwen3-8B} & Itemwise checklist and sentence-level genericity judge \\
Judge decoding & temperature 0.0 & Deterministic reward evaluation \\
Concurrency & 2048 & Maximum asynchronous reward-judge concurrency \\
Itemwise reward & true & Enables itemwise $R_{\mathrm{cov}}$ and $R_{\mathrm{mid}}$ \\
Rubric weighting & true & Uses rubric item weights in $R_{\mathrm{cov}}$ and $R_{\mathrm{mid}}$ \\
\bottomrule
\end{tabular}
\label{tab:reward_hparams}
\end{table}

\subsection{Inference-time length handling}
At inference, $|\mathcal{R}(x)|$ is unavailable. The reported run uses the generation budget above, with validation decoding set to temperature 0.6, top-$p$ 0.95, and one sample per prompt.

\section{Evaluation protocol details}
\label{app:evaluation_details}

For the reported \pager{} run, validation decoding uses temperature 0.6, top-$p$ 0.95, and one completion per prompt. The maximum prompt length is 6144 tokens and the maximum response length is 8192 tokens. All verifier and reward-judge calls use deterministic decoding with temperature 0.0.

CAPTURE, DCScore-F1, facts per 100 tokens, and average length are computed from saved benchmark predictions. $\mathrm{CapArena}_4$ is evaluated on IIW-400 and DOCCI using the four methods displayed in Table~\ref{tab:global_results}. Each displayed baseline is compared directly against \pager{}, and each judgment returns one of three outcomes: \pager{} wins, the baseline wins, or tie. The reported $\mathrm{CapArena}_4$ values exclude ties from the denominator; the \pager{} row pools the three direct comparisons against the displayed baselines for each dataset.

The verifier reliability study contains 347 human-labeled rubric-level judgments. We evaluate LLM judges against human-majority labels and report Cohen's $\kappa$ and class-wise F1 in Table~\ref{tab:verifier_reliability}. The paper authors annotated these 347 held-out rubric-level judgments. The additional volunteer sentence audits and caption-preference study are described separately in Appendix~\ref{app:human_protocol}.
\section{Existing assets, licenses, and terms of use}
\label{app:asset_licenses}

Table~\ref{tab:asset_licenses} summarizes the existing assets used in this work, their role in the paper, and the licenses or terms of use identified from the corresponding public project pages or model/data cards. We use all assets for research and evaluation purposes, cite the original creators, and do not redistribute third-party model weights, images, datasets, or evaluation resources unless their licenses explicitly permit redistribution.

\begin{table}[h]
\centering
\small
\setlength{\tabcolsep}{3pt}
\caption{Existing assets used in this work and their licenses or terms of use. ``No redistribution'' means that the paper reports aggregate results or uses the resource internally, but does not repackage or release the third-party asset.}
\begin{tabular}{>{\raggedright\arraybackslash}p{0.20\linewidth}>{\raggedright\arraybackslash}p{0.22\linewidth}>{\raggedright\arraybackslash}p{0.23\linewidth}>{\raggedright\arraybackslash}p{0.26\linewidth}}
\toprule
Asset & Role in this paper & License / terms identified & Use in this work \\
\midrule
Qwen3-VL-4B-Instruct~\citep{bai2025qwen3}
& Base captioning model and main GRPO initialization
& Apache-2.0
& Used for research training and evaluation; no third-party weights are redistributed by this paper. \\
\midrule
Qwen3-VL-8B / Qwen3-8B~\citep{bai2025qwen3}
& Evaluated captioner and checklist / reward judge
& Apache-2.0
& Used for deterministic judging and evaluation; no third-party weights are redistributed by this paper. \\
\midrule
InternVL3.5-8B / InternVL3.5-4B~\citep{wang2025internvl3}
& Evaluated captioners in the diagnostic study
& Apache-2.0
& Used for research evaluation; no third-party weights are redistributed by this paper. \\
\midrule
LLaVA-v1.5-7B~\citep{liu2023visual}
& Evaluated captioner in the diagnostic study
& Apache-2.0 for the project; use is also subject to the licenses and terms of its underlying datasets and base models
& Used for research evaluation only; no LLaVA weights, datasets, or checkpoints are redistributed. \\
\midrule
DeepSeek-V3.2~\citep{liu2025deepseek}
& Independent verifier in the reliability study
& MIT
& Used for verifier reliability and independent diagnostic comparison; no model weights are redistributed. \\
\midrule
PixMo-Cap / PixMo-derived captions~\citep{deitke2025molmo}
& Source of long-caption training data for the final GRPO run
& ODC-BY-1.0; intended for research and educational use under Ai2 Responsible Use Guidelines; generated text may also be subject to provider terms
& Used to construct filtered training prompts/rubrics; derived training data are not redistributed unless allowed by the upstream terms. \\
\midrule
DOCCI~\citep{onoe2024docci}
& Dense-caption evaluation and rubric construction source
& CC BY 4.0
& Used for research evaluation and rubric construction with attribution. \\
\midrule
DCI / densely captioned images~\citep{urbanek2024picture}
& Dense-caption data used in rubric construction and validation
& CC-BY-NC; underlying images are from SA-1B and require compliance with the SA-1B license agreement
& Used for non-commercial research; underlying images are not redistributed. \\
\midrule
ImageInWords / IIW-400~\citep{garg2024imageinwords}
& Main dense-caption evaluation benchmark
& CC BY 4.0
& Used for benchmark evaluation with attribution. \\
\midrule
CAPTURE~\citep{dong2024benchmarking}
& Global caption-quality metric
& Apache-2.0
& Used for evaluation; metric outputs are reported in aggregate. \\
\bottomrule
\end{tabular}
\label{tab:asset_licenses}
\end{table}

\paragraph{Additional assets.}
The text-only Qwen3-8B/14B judges and Qwen3-30B-A3B-Thinking extractor are distinct from the Qwen3-VL policies; the Qwen text-model family is described in \citet{yang2025qwen3text}. Additional 8B/32B policies use the Qwen3-VL family. GPT-5.5, Gemini-3.1-Pro, and Qwen-Plus are hosted-service labels governed by provider terms, rather than inferred open-weight licenses. No third-party model weights or dataset packages are redistributed with this manuscript.

\clearpage
\section{Qualitative case study}
\label{app:qualitative}

\newcommand{\corr}[1]{\textcolor{green!45!black}{\textbf{#1}}}
\newcommand{\hall}[1]{\textcolor{red!70!black}{\textbf{#1}}}
\newcommand{\irrelevant}[1]{\textcolor{gray!70!black}{\textit{#1}}}

\begin{figure}[t]
\centering
\includegraphics[width=0.58\linewidth]{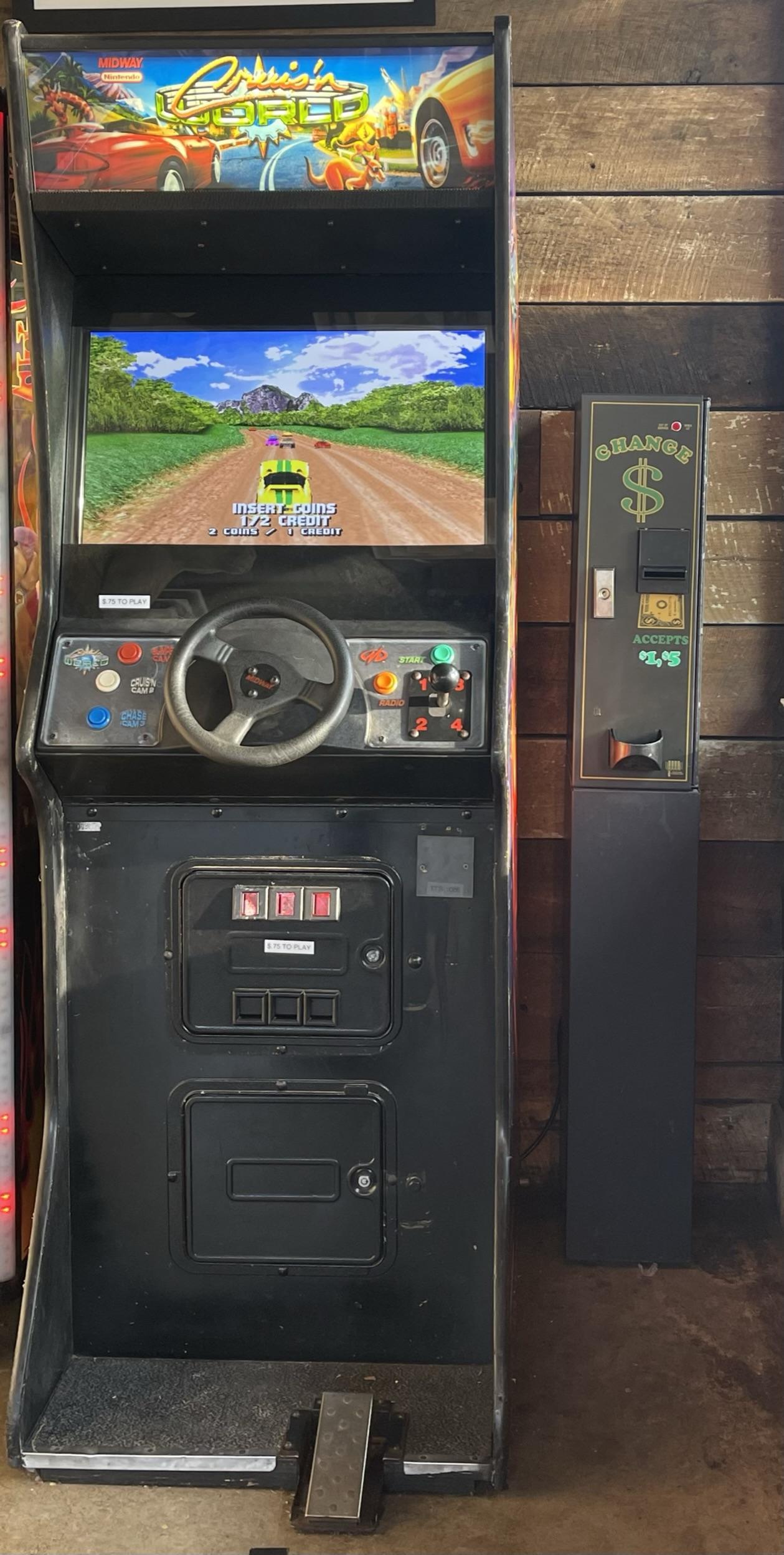}
\caption{\textbf{Case-study image \texttt{aar\_test\_04938}.}
The image shows an old \textit{Cruis'n World} arcade racing cabinet standing
next to a black change machine against a horizontal wooden plank wall.}
\label{fig:case_04938}
\end{figure}

Figure~\ref{fig:case_04938} shows a representative qualitative example for
image \texttt{aar\_test\_04938}. We include the image, the dense human reference,
and captions from Qwen3-VL-4B-Instruct, CapRL-Qwen3VL-4B, and \pager{}.
Colored spans mark representative local judgments:
\corr{green} indicates visually supported content, \hall{red} indicates
incorrect or unsupported specific content, and \irrelevant{gray} indicates
generic or non-evidential content. This case is illustrative; the quantitative
claims are supported by the aggregate results in
Tables~\ref{tab:main_position_results}--\ref{tab:ablations}.

\begin{table}[t]
\centering
\small
\setlength{\tabcolsep}{4pt}
\caption{\textbf{Summary of local caption behavior for
Figure~\ref{fig:case_04938}.} The base model is mostly correct at the global
level but ends with generic interpretation. CapRL is long and structured, but
introduces a salient title error and speculative game identification. \pager{}
keeps more of the caption anchored to visible text, hardware, layout, and
environmental evidence.}
\begin{tabular}{>{\raggedright\arraybackslash}p{0.17\linewidth}>{\raggedright\arraybackslash}p{0.24\linewidth}>{\raggedright\arraybackslash}p{0.24\linewidth}>{\raggedright\arraybackslash}p{0.25\linewidth}}
\toprule
Method & Supported evidence & Hallucinated / unsupported detail & Generic or non-evidential content \\
\midrule
Qwen3-VL-4B-Instruct
&
Correctly identifies a \textit{Cruis'n World} racing cabinet, active racing
screen, steering wheel, change machine, wooden wall, and concrete-like floor.
&
Introduces a ``dragon'' in the marquee and misreads the change machine as
accepting ``\$1.50''.
&
Ends with broad nostalgic framing such as ``ready for play'' and ``retro
atmosphere''.
\\
\midrule
CapRL-Qwen3VL-4B
&
Covers the arcade cabinet, screen, steering wheel, coin area, foot pedal, change
machine, wooden wall, and left-edge red lights.
&
Misreads the title as \textit{Crashin' World} and speculates that the game is a
version of \textit{Crash Bandicoot}.
&
Uses verbose sectioning and late explanatory language that is not directly
visual evidence.
\\
\midrule
\textbf{\pager{}}
&
Correctly identifies the \textit{Cruis'n World} cabinet, screen text, control
layout, coin hardware, change machine, wooden wall, floor, and left-edge red
LEDs.
&
Minor risk remains in over-specific decorative marquee interpretation, but the
caption avoids the major title error and unrelated franchise speculation.
&
Contains a short meta-style ending, but most tail content remains grounded in
visible layout, text, materials, and colors.
\\
\bottomrule
\end{tabular}
\label{tab:case_04938_summary}
\end{table}

\subsection{Dense Human Reference}
\label{app:case_gt_04938}

\begin{quote}
\normalsize
A \corr{narrow portrait-oriented indoor shot} of an old
\corr{\textit{Cruise'n} standup arcade game} placed alongside an
\corr{arcade change machine} inside a building with
\corr{horizontal wood plank walls}. The arcade game is very tall and narrow,
with a top rectangle that colorfully portrays the game title, along with an
\corr{orange kangaroo} in the middle of a \corr{red car racing in from the left}
and an \corr{orange car racing in from the right}. The cartoon road itself is
under the word \corr{``Cruise'n'' displayed in cursive handwriting in yellow},
and the word \corr{``WORLD,'' printed in green block letters against a blue sky}.
The action monitor for the arcade game is directly below that top rectangle.
The monitor displays a \corr{yellow cartoon race car driving away from the
camera} on a \corr{wide dirt track} where \corr{several other cars are ahead of
it}, and \corr{green grass and tall trees} appear on either side under a
\corr{blue sky with white clouds}. Partially covering the yellow car's rear
bumper are white words that read \corr{``INSERT COINS''} stacked on top of
\corr{``1/2 CREDITS''}, with indescribable text stacked below.

Below the monitor is a simple \corr{black plastic steering wheel} on a
\corr{black plastic dashboard with lights in red, white, blue, orange, and
green}. Below the steering wheel is a \corr{long black rectangle with a place to
put money into the machine} and a place to take out money once a square black
door is unlocked. Finally, on the floor of the game, partly resting on the
floor of the building, is the \corr{accelerator pedal}, probably made of metal
and brown in color. The floor of the building that houses the arcade game is
\corr{tan and stained}, the back wall is \corr{brown}, and a \corr{tall, thin
black change machine with a green dollar sign and the word ``CHANGE'' arched
over it} stands at the back wall.
\end{quote}

\subsection{Model Captions with Local Annotations}
\label{app:case_captions_04938}

\paragraph{CapRL-Qwen3VL-4B.}

\begin{quote}
\normalsize
\irrelevant{Based on the provided image, here is a detailed description:}

\irrelevant{\textbf{Image Caption:}}

The image shows a \corr{vintage arcade game machine}, specifically a racing
game titled \hall{``Crashin' World,''} standing next to a \corr{change machine}
against a \corr{wooden plank wall}.

\irrelevant{\textbf{Detailed Breakdown:}}

\irrelevant{\textbf{The Arcade Machine: ``Crashin' World''}}

\begin{itemize}
\item \textbf{Game Title:} The top marquee displays \hall{``Crashin' World''}
in a stylized, colorful font. The \corr{``MIDWAY Nintendo'' logo} is visible
in the top-left corner. The artwork features a \corr{red sports car}, a
\corr{kangaroo}, and a \hall{yellow car} on a dirt road.

\item \textbf{Screen:} The screen is active and displays a \corr{racing game in
progress}. A \corr{bright yellow-green car is driving down a dirt road} under a
\corr{blue sky with green hills and mountains}. The on-screen text reads:
\corr{``INSERT COINS 1/2 CREDIT''} and \corr{``2 COINS / 1 CREDIT.''}

\item \textbf{Control Panel:} Below the screen is a \corr{black, retro-style
control panel}.
\begin{itemize}
\item A \corr{large, silver or black steering wheel} is in the center, with a
\corr{``MIDWAY'' logo} in the middle.
\item To the left of the wheel are three buttons: a white button labeled
\hall{``CRUSH CAM,''} a red button, and a blue button labeled
\corr{``CHASE CAM.''}
\item To the right is a \corr{``START'' button} and a \corr{``RADIO'' button}
with a black joystick or selector knob labeled \corr{1, 2, 3, 4}.
\item A small white sticker reads \corr{``\$0.75 TO PLAY''} next to the
steering wheel.
\end{itemize}

\item \textbf{Coin Mechanism:} Below the steering wheel is a \corr{black panel
with three red, illuminated slots}, likely for coin insertion, with a
\corr{``\$0.75 TO PLAY'' sticker} below them. Beneath that is a
\corr{black rectangular door or compartment}.

\item \textbf{Foot Pedal:} At the bottom, a \corr{metal foot pedal} is visible,
likely for the accelerator, with a textured, diamond-plate pattern.

\item \textbf{Machine Color:} The entire machine is \corr{black}, with a worn,
slightly dirty appearance, suggesting age or use.
\end{itemize}

\textbf{Game in Progress:} The game appears to be \hall{``Crashin' World,''}
a racing game from Midway, \hall{likely a version of ``Crash Bandicoot'' or a
similar racing title}, as it features a kangaroo and a dirt track.

\textbf{Change Machine:}

\begin{itemize}
\item To the right of the arcade machine is a \corr{black, vertical change
machine}.
\item It has a \corr{green dollar sign} and the word \corr{``CHANGE''} at the
top.
\item Below it, it says \hall{``ACCEPTS \$1.5''} in green text, next to a slot
for a dollar bill or coin.
\item It has a small slot with a dollar bill graphic and a metal coin slot
below it.
\end{itemize}

\textbf{Background:}

\begin{itemize}
\item The arcade machine is set against a wall of \corr{dark, weathered wooden
planks}.
\item The floor appears to be \corr{concrete} or \hall{a dirt floor}, with a
brownish, earthy tone.
\item A \corr{vertical strip of red LED lights} is visible on the far left,
likely from another arcade machine.
\end{itemize}
\end{quote}

\paragraph{Qwen3-VL-4B-Instruct.}

\begin{quote}
\normalsize
This image shows a \corr{vintage arcade game machine}, specifically
\corr{``Cruis'n World,''} standing in a room with a \corr{rustic wooden wall}.
The machine is a \corr{black cabinet} with a \corr{large screen displaying the
game in progress}.

\textbf{Main Game Cabinet:}
\begin{itemize}
\item \textbf{Top Marquee:} The top of the cabinet features a colorful, vibrant
marquee with the game's title, \corr{``Cruis'n World,''} in stylized yellow
and green lettering. The logo includes the \corr{``Midway'' and ``Nintendo''
brand names}. The artwork depicts a \corr{red sports car} racing on a road
with a \corr{kangaroo} and a \hall{dragon} in a tropical, cartoonish landscape.

\item \textbf{Screen:} The screen is active, showing a \corr{3D racing game
scene}. A \corr{yellow race car is driving down a dirt road} with
\corr{green hills and a blue sky}. Text on the screen reads
\corr{``INSERT COINS 1/2 CREDIT''} and \corr{``2 COINS / 1 CREDIT.''}

\item \textbf{Control Panel:} Below the screen is the control panel, which
includes:
\begin{itemize}
\item A \corr{large, black, three-spoke steering wheel}.
\item A set of buttons labeled \corr{``CHASE CAM''} and \hall{``CRUSH CAM''}
on the left.
\item A \corr{``START'' button} with a green light, a \corr{``RADIO'' button}
with an orange light, and a selector knob with numbers \corr{1, 2, 3, and 4}.
\item A small, rectangular panel with \corr{three red lights}, likely indicating
the game's status or credits.
\item A small, rectangular panel with a label that reads \corr{``\$.75 TO
PLAY.''}
\end{itemize}

\item \textbf{Foot Pedal:} At the bottom of the cabinet, there is a
\corr{metal foot pedal} for controlling the car's acceleration.
\end{itemize}

\textbf{Change Machine:}
\begin{itemize}
\item To the right of the arcade machine is a \corr{tall, black change machine}
with a \corr{green dollar sign symbol} and the word \corr{``CHANGE''} at the top.
It has a slot for inserting coins or bills and a small display indicating it
accepts \hall{``\$1.50.''}
\end{itemize}

\textbf{Background:}
\begin{itemize}
\item The arcade machine is positioned against a wall made of
\corr{dark, weathered wooden planks}, giving the setting a
\irrelevant{rustic or garage-like feel}.
\item The floor is \corr{concrete}, and the overall lighting is
\hall{dim}, with some ambient light illuminating the machine.
\end{itemize}

\irrelevant{The image captures a nostalgic moment of arcade gaming, with the
machine ready for play and the surrounding environment adding to the retro
atmosphere.}
\end{quote}

\paragraph{\pager{}.}

\begin{quote}
\normalsize
This is a \corr{full-frontal, eye-level photograph} of a vintage
\corr{``Cruis'n World'' arcade racing game cabinet} by Midway, licensed from
Nintendo, positioned indoors against a \corr{rustic wooden plank wall}. To its
right stands a \corr{tall, black change machine}.

\textbf{Arcade Cabinet -- Upper Section:}
\begin{itemize}
\item \textbf{Marquee:} At the top, a colorful illuminated marquee displays the
game's title \corr{``Cruis'n World''} in stylized yellow and green lettering
with a metallic sheen. The background artwork features \corr{red sports cars}
speeding on a road, flanked by tropical scenery including \hall{palm trees,
mountains, and cartoonish dinosaurs}. Logos for \corr{``MIDWAY'' and
``Nintendo''} appear in the upper left corner.

\item \textbf{Screen:} Below the marquee, a CRT monitor shows active gameplay:
a third-person view of a \corr{bright yellow car driving down a dirt road}
through a \corr{lush green landscape} under a \corr{blue sky with white clouds}.
\corr{Other colored cars are visible ahead}. On-screen text reads:
\corr{INSERT COINS}, \corr{1/2 CREDIT}, and \corr{2 COINS / 1 CREDIT}. A small
white sticker to the left of the screen says \corr{``\$ .75 TO PLAY''}.
\end{itemize}

\textbf{Control Panel:}
\begin{itemize}
\item Centered below the screen is a \corr{large, three-spoke steering wheel}
with a silver rim and black center hub bearing the \corr{``Midway'' logo}.
\item To the left of the wheel: four circular buttons --- red
\corr{``CRUIS'N CAM \#''}, white, blue \corr{``CHASE CAM 3''}, and another
unlabeled button above them. Above these is a small decal reading \corr{``USA''}.
\item To the right of the wheel: an orange \corr{``RADIO'' button}, a green
\corr{``START'' button}, and a black gear shifter with positions labeled
\corr{1--4} in red numerals around it.
\item The entire control panel surface is \corr{dark gray or black plastic}
with slight wear.
\end{itemize}

\textbf{Lower Cabinet Body:}
\begin{itemize}
\item The main body is \corr{matte black metal/plastic} with visible screws
and seams.
\item Two recessed panels are stacked vertically:
\begin{itemize}
\item \textbf{Upper Panel:} Contains \corr{three vertical coin slots with red
indicator lights} above each. Below them is a horizontal slot labeled
\corr{``\$ .75 TO PLAY''}, followed by three empty rectangular cutouts. A small
round keyhole is to the right.
\item \textbf{Lower Panel:} A larger access door with a horizontal handle and a
keyhole on the right side.
\end{itemize}
\item At the bottom front, a textured black rubber mat covers the floor area
where players stand.
\item Protruding forward from the base is a \corr{single, wide, metallic foot
pedal} with raised dots for grip, mounted on a black bracket.
\end{itemize}

\textbf{Change Machine (Right Side):}
\begin{itemize}
\item \corr{Tall, narrow, black unit} standing adjacent to the arcade cabinet.
\item Top section has gold-trimmed labeling: \corr{``CHANGE''} arched over a
large \corr{green dollar sign symbol}.
\item Below that: a bill acceptor slot, then a graphic of a \$1 bill, and text
reading \corr{``ACCEPTS \$1, \$5''} in green.
\item Further down: a curved metal tray for dispensing coins.
\item Small red LED indicator light near the top right.
\item The lower half is plain black with no controls or labels.
\end{itemize}

\textbf{Background \& Environment:}
\begin{itemize}
\item Wall behind both machines consists of \corr{horizontally aligned,
weathered wood planks} in varying shades of brown, showing grain, knots, and
gaps between boards.
\item Floor appears to be \corr{concrete or painted cement}, slightly stained
and worn.
\item Lighting is ambient indoor illumination, casting soft shadows beneath the
machines.
\item Partially visible to the far left edge is another arcade machine with
\corr{glowing red vertical LEDs} along its side.
\end{itemize}

\textbf{Textures \& Colors Summary:}
\begin{itemize}
\item Arcade cabinet: predominantly \corr{black} with glossy screen, matte
control panel, metallic accents on wheel and pedal.
\item Marquee: vibrant blues, greens, reds, yellows.
\item Change machine: \corr{matte black} with gold trim and green/red
text/symbols.
\item Wood wall: \corr{rough, natural texture with earthy browns}.
\item Overall impression: \irrelevant{well-used but functional retro gaming
equipment in a casual or bar-like setting}.
\end{itemize}

\corr{No people are present}. All visible text is legible and transcribed as
shown. \irrelevant{No speculative details about functionality beyond what is
visibly labeled or implied by standard arcade design.}
\end{quote}

\paragraph{Case-study takeaway.}
This example illustrates differences in the distribution of detail and highlights both grounded descriptions and remaining errors. The base model ends with generic atmosphere and includes specific errors. CapRL's incorrect title reading, \hall{\textit{Crashin' World}}, leads to further speculation about \hall{\textit{Crash Bandicoot}}. More of \pager{}'s late description refers to visible objects and layout, but its annotated errors remain. The color annotations show image-based assessments; the quantitative diagnostic uses reference-induced labels.
\section{Diagnostic limitations}
\label{app:limitations}
The diagnostic measures agreement with a dense reference, rather than exhaustive image entailment. A visually correct detail absent from the reference can remain \generic{}; a false \contra{} label requires active conflict with an imperfect reference. Appendix~\ref{app:image_audits} checks supported and contradicted content against images; Appendix~\ref{app:post_rl} describes residual-label composition after training. The sampled image recheck motivates an additional confirmation step for contradictions, but the reported main experiments retain the original rule. Reference omissions, ambiguity, and shared judge biases remain relevant limitations.

\section{Broader impacts}
\label{app:broader_impacts}

Detailed image captioning can have positive societal impacts in accessibility, assistive technology, image search, dataset auditing, and visual-content understanding. By diagnosing where long captions become unsupported or generic, and by training models to maintain evidence-bearing detail, this work may help make generated visual descriptions more faithful and easier to audit.

The same capabilities can also create risks. More fluent and detailed captions may increase user over-trust, especially when a caption is used in assistive, educational, journalistic, or safety-relevant settings. Captioning systems can also expose sensitive information present in images, support large-scale surveillance or profiling, or reproduce dataset and verifier biases about people, places, activities, and cultural contexts. In addition, our reference-supported diagnostic may mark visually correct but reference-omitted details as unsupported, which can create uneven behavior across image domains.

We do not deploy the system in a real-world application, do not evaluate it for high-stakes decision-making, and do not release a high-risk scraped dataset. Responsible use of detailed captioning systems should include privacy filtering, clear communication of uncertainty, human review in high-stakes settings, compliance with dataset and model licenses, and additional evaluation on demographic, cultural, and domain-specific failure modes before deployment.
\clearpage
\section{Additional validation and robustness experiments}
\label{app:additional_validation}

This appendix provides complementary validation through image-based audits, human preferences, post-training residual-content analysis, an independent-verifier comparison, and stronger-model diagnosis and training. These studies examine visual validity, caption preference, judge consistency, and transfer under the tested settings.

\subsection{Human evaluation protocol}
\label{app:human_protocol}

The additional human evaluations used unpaid volunteers recruited from the authors' personal circle, including friends. Participation was voluntary. System identities were hidden, but personal familiarity with the authors remains a potential source of bias; the sample is not a representative panel of users.

Two annotators conducted the image-based and post-training residual-label audits. In the image-based audits, they inspected the source image and classified selected claims or sentences as visually confirmed, incorrect, or ambiguous. The residual-label audit distinguished filler, correct unreferenced details, and visually incorrect details. Source systems and study hypotheses were hidden for these audits.

Five volunteers performed the preference study, ranking three anonymized captions per image in randomized order and providing a short justification. The study covers 100 images and 500 first-place judgments, with ratings clustered by image and annotator. The original 347-item verifier study in Appendix~\ref{app:verifier} used author annotations and is distinct from these additional volunteer studies.

\paragraph{Ethics and participant protection.}
The research conforms to the NeurIPS Code of Ethics. Potential risks associated with the evaluation tasks were considered and disclosed to participants. The studies followed the applicable institutional ethics review requirements, and the required review or equivalent approval was obtained. No monetary compensation was provided.

\subsection{Image-based audits}
\label{app:image_audits}

Two volunteers checked 150 \supp{} claims from \pager{}, sampled across rubric dimensions and caption quartiles: 142 (94.7\%) were confirmed against the image and 8 (5.3\%) were unconfirmed or ambiguous, with inter-annotator $\kappa=.88$. The unconfirmed category includes visually ambiguous cases.

A separate audit examined 150 \contra{} sentences across the six diagnosed models and four quartiles: 139 (92.7\%) conflicted with the image, 3 (2.0\%) were correct but conflicted with the reference, and 8 (5.3\%) were ambiguous or partially occluded, with inter-annotator $\kappa=.87$.

\begin{table}[htbp]
\centering\small
\caption{Image-based audits of supported claims from PAGER and contradicted sentences from the six diagnosed models. Results are descriptive.}
\begin{tabular}{p{.21\linewidth}p{.47\linewidth}rr}
\toprule
Audit & Outcome & Count & Share \\
\midrule
\supp{} claims & Confirmed true & 142/150 & 94.7\% \\
(\pager{}) & Not confirmed / ambiguous & 8/150 & 5.3\% \\
\midrule
\contra{} sentences & Genuine image conflict & 139/150 & 92.7\% \\
(six models) & Correct but conflicts with reference & 3/150 & 2.0\% \\
 & Ambiguous / partly occluded & 8/150 & 5.3\% \\
\bottomrule
\end{tabular}
\label{tab:image_audits}
\end{table}

Relabeling the three false contradictions changed the six-model mean $D_{\mathrm{mid}}$ from $.101$ to $.099$ (a change below $.005$). This sampled recheck motivates retaining a \contra{} label only after image confirmation. The recheck concerns the sampled claims; the main experiments use the original reference-supported rule. These cases identify reference errors as a source of pseudo-contradictions; the sensitivity result applies to the sampled recheck.

\subsection{Blind three-way human preference}
\label{app:human_preference}

Table~\ref{tab:human_preference} summarizes the first-place votes: 389/500 (77.8\%) for \pager{}, with every rater's overall ordering \pager{} $>$ CapRL $>$ SFT. Fleiss' $\kappa=.66$ summarizes agreement across the three first-choice categories. Shares use the three-way first-choice vote denominator.

\begin{table}[htbp]
\centering\small
\caption{Blind preference study with five unpaid volunteers: 100 images and 500 first-place votes.}
\begin{tabular}{lrrc}
\toprule
Method & Votes & Share & Per-rater share range \\
\midrule
\pager{} & 389 & 77.8\% & 72--84\% \\
CapRL & 91 & 18.2\% & 14--22\% \\
SFT & 20 & 4.0\% & 2--6\% \\
\bottomrule
\end{tabular}
\label{tab:human_preference}
\end{table}

The raters' explanations mentioned anchored transitions and fuller coverage for \pager{}, occasional formulaic phrasing, list-like descriptions for CapRL, and omissions for SFT. These comments qualitatively support the observed caption-level preferences and positional patterns.

\subsection{Post-training residual content and lexical diversity}
\label{app:post_rl}

The post-training audit sampled about 100 qualifying $Q_4$ \generic{} sentences per condition from Base and \pager{} on the same 198-image set. The candidate pools were approximately 210 and 130 sentences, respectively. Two volunteers applied the same image-based three-way classification, blind to the system. The audit measures composition within the selected residual-sentence class.

\begin{table}[htbp]
\centering\small
\caption{Post-RL residual-label percentages and differences (percentage points), with 95\% bootstrap intervals. The comparison is descriptive.}
\begin{tabular}{p{.36\linewidth}rrc}
\toprule
Category & Base (\%) & \pager{} (\%) & Difference [95\% CI] \\
\midrule
Non-evidential filler & 88 & 81 & $-7.0\;[-17.4,3.6]$ \\
Correct, absent from reference & 8 & 14 & $+6.0\;[-2.8,15.1]$ \\
Visually incorrect & 4 & 5 & $+1.0\;[-5.2,7.3]$ \\
\midrule
Inter-annotator $\kappa$ & .84 & .85 & --- \\
\bottomrule
\end{tabular}
\label{tab:post_rl_generic}
\end{table}

All three intervals include zero, so these differences are interpreted descriptively. Percentages are conditional on the selected residual class; changes in its size also affect the corresponding absolute counts.

The diversity comparison uses two $Q_4$ sentences per caption from the same 198 images. The comparison controls sentence count while retaining each system's generated length; overall captions averaged 257 tokens for Base and 412 for \pager{}. The changes in Table~\ref{tab:tail_diversity} indicate a modest stylistic concentration and are consistent with the occasional formulaic phrasing noted by human raters.

\begin{table}[htbp]
\centering\small
\caption{Tail lexical diversity with two sentences per caption.}
\begin{tabular}{lcc}
\toprule
Metric & Base & \pager{} \\
\midrule
Distinct-1 & .71 & .68 \\
Distinct-2 & .58 & .55 \\
Self-BLEU-4 & .19 & .24 \\
\bottomrule
\end{tabular}
\label{tab:tail_diversity}
\end{table}

During reward development, coverage-only optimization produced repeated or reworded high-weight details. Length regularization addressed this behavior; a fixed budget could then over-compress information-rich images, motivating the rubric-dependent budget. These qualitative development observations motivated the content-aware length regularizer.

\subsection{Independent-verifier comparison}
\label{app:cross_judge}

Qwen3-8B supplies training and main diagnostic judgments. DeepSeek-V3.2 never enters the GRPO loop. On the 347 human-labeled rubric judgments, their Cohen's $\kappa$ values are $.89$ and $.92$, respectively (Appendix~\ref{app:verifier}). This reliability study is distinct from the full policy comparison below.

\begin{table}[htbp]
\centering\small
\caption{Position metrics under Qwen3-8B / DeepSeek-V3.2 on the same mixed IIW-400 + DOCCI diagnostic subset ($n=198$).}
\begin{tabular}{lcccc}
\toprule
Method & $D_{\mathrm{mid}}$ & $S_{Q_4}$ & $H_{Q_4}$ & $G_{Q_4}$ \\
\midrule
Base & .096 / .089 & .290 / .301 & .298 / .289 & .412 / .410 \\
CapRL & .125 / .118 & .395 / .388 & .283 / .276 & .322 / .336 \\
\pager{} & .036 / .041 & .436 / .429 & .235 / .241 & .329 / .330 \\
\bottomrule
\end{tabular}
\label{tab:cross_judge}
\end{table}

The rankings favor \pager{} on $D_{\mathrm{mid}}$, $S_{Q_4}$, and $H_{Q_4}$ under both judges. Residual genericity is slightly lower for CapRL under Qwen3-8B ($.322$ versus $.329$), but for \pager{} under DeepSeek-V3.2 ($.330$ versus $.336$). The diagnostic outputs have 90\% agreement and $\kappa=.83$. These checks provide evidence beyond the training judge, while two reference-based judges can still share systematic blind spots. The separate image audits and human preferences address complementary aspects of that risk.

\subsection{Stronger foundation-model diagnosis}
\label{app:foundation}

We applied the same reference-relative diagnostic, without retraining, to GPT-5.5 and Gemini-3.1-Pro on the same 198-image set. Qwen-Plus and the original six-model mean are included for context.

\begin{table}[htbp]
\centering\small
\caption{Additional foundation-model tail diagnostics.}
\begin{tabular}{lcc}
\toprule
Model / summary & $D_{\mathrm{tail}}$ & Reported 95\% CI \\
\midrule
Six-model mean (original) & .178 & --- \\
Qwen-Plus (original) & .138 & [.110,.169] \\
GPT-5.5 & .132 & [.101,.159] \\
Gemini-3.1-Pro & .124 & [.091,.154] \\
\bottomrule
\end{tabular}
\label{tab:foundation_tail}
\end{table}

Both additional models have positive tail effects within the original six-model range: InternVL3.5-8B has $.090$, Qwen3-VL-8B has $.122$ $[.098,.149]$, and LLaVA-v1.5-7B has $.380$. Thus the observed tail pattern is not confined to the smallest captioners. This comparison evaluates the positional diagnostic on pretrained foundation models.

\subsection{Short-budget transfer and three-seed replication}
\label{app:scaling}

The exploratory transfer study evaluated the same mixed IIW-400 + DOCCI diagnostic subset ($n=198$), with one 100-step run for each new variant: CapRL+\pager{} continuation and \pager{} initialized from Qwen3-VL-8B/32B. Each had a single evaluation pass; per-scale learning rates were adjusted, with the other main-run settings unchanged. The pretrained 8B baseline uses captions regenerated for this study and evaluated separately from the original diagnostic in Table~\ref{tab:unweighted_position_stats}. Table~\ref{tab:transfer_single} reports all seven metrics; generic rates satisfy $G_q=1-S_q-H_q$, and their shifts satisfy $D_{\mathrm{tail}}=-\Delta S-\Delta H$. The main 4B comparison is reproduced for context.

\begin{table}[htbp]
\centering\small
\caption{Short-budget transfer results. New trained variants use 100 steps. The two blocks contain the same rows; $\Delta S=S_{Q_4}-S_{Q_1}$ and $\Delta H=H_{Q_4}-H_{Q_1}$.}
\begin{tabular}{llcccc}
\toprule
Method & Backbone & $D_{\mathrm{mid}}$ & $S_{Q_4}$ & $H_{Q_4}$ & $G_{Q_4}$ \\
\midrule
Pretrained & 4B & .096 & .290 & .298 & .412 \\
Pretrained & 8B & .101 & .312 & .293 & .395 \\
Pretrained & 32B & .112 & .326 & .288 & .386 \\
CapRL & 4B & .125 & .395 & .283 & .322 \\
CapRL + \pager{} & 4B & .121 & .401 & .277 & .322 \\
\pager{} (main) & 4B & .036 & .436 & .235 & .329 \\
\pager{} & 8B & .042 & .455 & .221 & .324 \\
\pager{} & 32B & .044 & .459 & .217 & .324 \\
\bottomrule
\end{tabular}

\medskip
\begin{tabular}{llrrr}
\toprule
Method & Backbone & $\Delta S$ & $\Delta H$ & $D_{\mathrm{tail}}$ \\
\midrule
Pretrained & 4B & $-.241$ & $-.028$ & .269 \\
Pretrained & 8B & $-.203$ & $+.002$ & .201 \\
Pretrained & 32B & $-.194$ & $+.014$ & .180 \\
CapRL & 4B & $-.151$ & $+.028$ & .123 \\
CapRL + \pager{} & 4B & $-.147$ & $+.016$ & .131 \\
\pager{} (main) & 4B & $-.094$ & $-.083$ & .177 \\
\pager{} & 8B & $-.086$ & $-.065$ & .151 \\
\pager{} & 32B & $-.081$ & $-.060$ & .141 \\
\bottomrule
\end{tabular}
\label{tab:transfer_single}
\end{table}

The summaries show lower $D_{\mathrm{mid}}$ and $H_{Q_4}$ and higher $S_{Q_4}$ after \pager{} training relative to the corresponding pretrained rows. CapRL continuation changes $D_{\mathrm{mid}}$ from $.125$ to $.121$ and $\Delta H$ from $+.028$ to $+.016$. These comparisons assess short-budget transfer under the specified per-scale training settings. The evaluation uses caption-quality and positional metrics; model size and learning rate vary together.

The 8B/32B experiments were subsequently repeated with three training seeds at the same 100-step budget and evaluated on the same 198-image subset. Table~\ref{tab:scaling_three_seed} presents three positional metrics with 95\% $t$ intervals over training seeds.

\begin{table}[htbp]
\centering\small
\caption{Three-seed, 100-step update: mean [95\% $t$ CI], $n=3$.}
\begin{tabular}{lccc}
\toprule
Backbone & $D_{\mathrm{mid}}$ & $S_{Q_4}$ & $H_{Q_4}$ \\
\midrule
8B & .041 [.039,.043] & .457 [.423,.491] & .216 [.214,.219] \\
32B & .043 [.041,.045] & .467 [.455,.480] & .214 [.214,.215] \\
\bottomrule
\end{tabular}
\label{tab:scaling_three_seed}
\end{table}

The three-seed means are close to the corresponding single-run values. The $S_{Q_4}$ intervals overlap, with greater variability for 8B. These replications provide additional support for the three reported positional metrics under the tested short-budget settings.

\subsection{Construction details and measured cost}
\label{app:construction_cost}

\paragraph{Data filtering.}
Starting from 500,000 PixMo-derived candidates, an image-quality screen and mean-claim-length constraint retain 494,596. The length constraint excludes rubric sets with mean claim length below approximately five words. One pre-inference rollout per candidate is then scored by the judge, retaining 297,839 after removing uninformative or unsolvable cases and already-saturated examples. This offline filtering is distinct from online filtering during GRPO. The approach follows rubric-based curation in \citet{huang2025rubricanchors}.

\paragraph{Mixed evidence.}
\label{app:mixed_sensitivity}
The sensitivity sample contained 187 of 11,714 sentences (1.6\%) with both supported and conflicting evidence. Reversing the \supp{}-precedence rule changed $D_{\mathrm{mid}}$ and $D_{\mathrm{tail}}$ by less than $.01$. This tests label precedence; the unweighted diagnostic in Appendix~\ref{app:unweighted} separately tests rubric-dimension weighting.

\paragraph{Token footprint.}
Rubric extraction is an offline, reusable step; verification is repeated per generated caption. Table~\ref{tab:token_cost} reports mean token counts. These token counts cover extraction and itemwise verification; online training additionally includes sentence-level tail judging. The main 4B run used 32 A100 GPUs for approximately two days (Appendix~\ref{app:training}).

\begin{table}[htbp]
\centering\small
\caption{Mean input/output token counts per call.}
\begin{tabular}{p{.45\linewidth}rrp{.20\linewidth}}
\toprule
Stage / model & Input & Output & Frequency \\
\midrule
Rubric extraction / Qwen3-30B-A3B-Thinking & 1186 & 423 & Once per image \\
Itemwise verification / Qwen3-8B & 548 & 596 & Per caption \\
\bottomrule
\end{tabular}
\label{tab:token_cost}
\end{table}

\clearpage
\section*{NeurIPS Paper Checklist}

\begin{enumerate}

\item {\bf Claims}
    \item[] Question: Do the main claims made in the abstract and introduction accurately reflect the paper's contributions and scope?
    \item[] Answer: \answerYes{}
    \item[] Justification: The abstract, introduction, and discussion distinguish reference-relative diagnostics from image truth. Claims about preference and larger backbones are limited to the volunteer sample and short-budget experiments in Appendix~\ref{app:additional_validation}; the conclusions are scoped to those evaluation settings.


\item {\bf Limitations}
    \item[] Question: Does the paper discuss the limitations of the work performed by the authors?
    \item[] Answer: \answerYes{}
    \item[] Justification: Section~\ref{sec:discussion} and Appendices~\ref{app:limitations} and~\ref{app:additional_validation} discuss reference omissions, false contradictions, finite human samples, personal-circle recruitment, judge dependence, and learning-rate confounding.


\item {\bf Theory assumptions and proofs}
    \item[] Question: For each theoretical result, does the paper provide the full set of assumptions and a complete (and correct) proof?
    \item[] Answer: \answerNA{}
    \item[] Justification: The paper does not present formal theoretical results such as theorems or lemmas. The equations define the diagnostic representation, effect sizes, and reward components rather than theoretical claims requiring proof.


\item {\bf Experimental result reproducibility}
    \item[] Question: Does the paper fully disclose all the information needed to reproduce the main experimental results of the paper to the extent that it affects the main claims and/or conclusions of the paper (regardless of whether the code and data are provided or not)?
    \item[] Answer: \answerYes{}
    \item[] Justification: Sections~\ref{sec:framework}--\ref{sec:pager} define the diagnostic quantities and training objective. Appendix~\ref{app:pipeline} gives rubric extraction, itemwise verification, sentence aggregation, the tail-judge decision rule, and bootstrap procedures. Appendix~\ref{app:training} specifies the main model, data sources, optimizer and learning rate, decoding settings, and reward hyperparameters; Appendix~\ref{app:evaluation_details} describes evaluation. Together these details provide an implementation-level description of the method and settings underlying the main results.


\item {\bf Open access to data and code}
    \item[] Question: Does the paper provide open access to the data and code, with sufficient instructions to faithfully reproduce the main experimental results, as described in supplemental material?
    \item[] Answer: \answerNo{}
    \item[] Justification: The method and experimental settings are documented in the paper. The training code, checkpoints, and prediction files are not publicly released with this manuscript.


\item {\bf Experimental setting/details}
    \item[] Question: Does the paper specify all the training and test details (e.g., data splits, hyperparameters, how they were chosen, type of optimizer) necessary to understand the results?
    \item[] Answer: \answerYes{}
    \item[] Justification: Section~\ref{sec:pager} defines the reward and optimization objective. Appendix~\ref{app:training}, including Tables~\ref{tab:training_hparams}--\ref{tab:reward_hparams}, specifies the main 4B model, data sources, optimizer, learning rate, training budget, decoding settings, and reward coefficients. Appendices~\ref{app:pipeline} and~\ref{app:evaluation_details} describe the diagnostic and evaluation protocols. Appendix~\ref{app:additional_validation} states the scope and settings of the supplementary studies.


\item {\bf Experiment statistical significance}
    \item[] Question: Does the paper report error bars suitably and correctly defined or other appropriate information about the statistical significance of the experiments?
    \item[] Answer: \answerYes{}
    \item[] Justification: The cross-model diagnostic uses 95\% image-level bootstrap CIs. Appendix~\ref{app:scaling} reports 95\% t intervals over three training seeds. The post-RL intervals span zero and are interpreted descriptively. Evaluation repeats, training seeds, and image-clustered human ratings are identified separately in the experimental protocols.


\item {\bf Experiments compute resources}
    \item[] Question: For each experiment, does the paper provide sufficient information on the computer resources (type of compute workers, memory, time of execution) needed to reproduce the experiments?
    \item[] Answer: \answerNo{}
    \item[] Justification: The compute disclosure covers the main 32-A100 run (approximately two days) and extraction/itemwise-verification token averages in Appendices~\ref{app:training} and~\ref{app:construction_cost}. These are component-level measurements rather than total project compute.


\item {\bf Code of ethics}
    \item[] Question: Does the research conducted in the paper conform, in every respect, with the NeurIPS Code of Ethics \url{https://neurips.cc/public/EthicsGuidelines}?
    \item[] Answer: \answerYes{}
    \item[] Justification: The research conforms to the NeurIPS Code of Ethics. The paper discloses the existing datasets and the additional volunteer evaluations, including personal-circle recruitment and the absence of monetary compensation. Appendix~\ref{app:human_protocol} describes the evaluation protocol and compliance with the applicable institutional ethics review requirements.


\item {\bf Broader impacts}
    \item[] Question: Does the paper discuss both potential positive societal impacts and negative societal impacts of the work performed?
    \item[] Answer: \answerYes{}
    \item[] Justification: Appendix~\ref{app:broader_impacts} discusses positive impacts such as accessibility, assistive technology, image search, dataset auditing, and more faithful visual descriptions. It also discusses risks including over-trust, privacy leakage, surveillance or profiling, dataset and verifier bias, and the need for safeguards such as privacy filtering, uncertainty communication, human review, and license compliance.


\item {\bf Safeguards}
    \item[] Question: Does the paper describe safeguards that have been put in place for responsible release of data or models that have a high risk for misuse (e.g., pre-trained language models, image generators, or scraped datasets)?
    \item[] Answer: \answerNA{}
    \item[] Justification: This manuscript does not describe a new public release of a high-risk scraped dataset or model checkpoint. Potential misuse and deployment safeguards are discussed in Appendix~\ref{app:broader_impacts}. Any later asset release requires its own documentation and safeguards.


\item {\bf Licenses for existing assets}
    \item[] Question: Are the creators or original owners of assets (e.g., code, data, models), used in the paper, properly credited and are the license and terms of use explicitly mentioned and properly respected?
    \item[] Answer: \answerNo{}
    \item[] Justification: Appendix~\ref{app:asset_licenses} credits the source assets and lists identified licenses and terms. Hosted models are governed by provider service terms; the appendix does not enumerate all version-specific terms.


\item {\bf New assets}
    \item[] Question: Are new assets introduced in the paper well documented and is the documentation provided alongside the assets?
    \item[] Answer: \answerNA{}
    \item[] Justification: This paper presents a method and experimental evaluations using existing assets; it does not accompany a new public dataset, code, or checkpoint release.


\item {\bf Crowdsourcing and research with human subjects}
    \item[] Question: For crowdsourcing experiments and research with human subjects, does the paper include the full text of instructions given to participants and screenshots, if applicable, as well as details about compensation (if any)? 
    \item[] Answer: \answerNo{}
    \item[] Justification: Appendix~\ref{app:human_protocol} describes the additional evaluations, task criteria, blinding, and recruitment of unpaid friends who volunteered to help. No monetary compensation was provided. The original instruction sheets and interface screenshots are not included. The original 347-item author-annotation study is distinguished from the additional volunteer studies.


\item {\bf Institutional review board (IRB) approvals or equivalent for research with human subjects}
    \item[] Question: Does the paper describe potential risks incurred by study participants, whether such risks were disclosed to the subjects, and whether Institutional Review Board (IRB) approvals (or an equivalent approval/review based on the requirements of your country or institution) were obtained?
    \item[] Answer: \answerYes{}
    \item[] Justification: Appendix~\ref{app:human_protocol} describes the volunteer evaluations and states that potential participant risks were considered and disclosed and that the applicable institutional ethics review requirements were followed, with the required review or equivalent approval obtained.


\item {\bf Declaration of LLM usage}
    \item[] Question: Does the paper describe the usage of LLMs if it is an important, original, or non-standard component of the core methods in this research? Note that if the LLM is used only for writing, editing, or formatting purposes and does \emph{not} impact the core methodology, scientific rigor, or originality of the research, declaration is not required.
    \item[] Answer: \answerYes{}
    \item[] Justification: LLMs are central to rubric extraction, itemwise verification, and tail-genericity judging. The paper distinguishes Qwen3-8B training/main judging from DeepSeek-V3.2 evaluation outside the GRPO loop, and reports the extractor and additional captioning models in Appendix~\ref{app:additional_validation}.


\end{enumerate}

\end{document}